\documentclass[conference]{IEEEtran}
\IEEEoverridecommandlockouts
\usepackage{cite}
\usepackage{amsmath,amssymb,amsfonts}
\usepackage{graphicx}
\usepackage{textcomp}
\def\BibTeX{{\rm B\kern-.05em{\sc i\kern-.025em b}\kern-.08em
    T\kern-.1667em\lower.7ex\hbox{E}\kern-.125emX}}

\usepackage{comment}
\usepackage{grffile}
\usepackage[justification=centering]{caption}
\usepackage{subcaption}
\usepackage[font=small,labelfont=bf]{caption}
\usepackage{float}
\usepackage{subfloat}
\usepackage{tabularx}
\usepackage{algorithm}
\usepackage[noend]{algpseudocode}
\usepackage{parskip}
\usepackage[super]{nth}		

\newcolumntype{M}[1]{>{\centering\arraybackslash}m{#1}}
\newcolumntype{N}{@{}m{0pt}@{}}

\usepackage{bibspacing}
\usepackage{amsmath, amsfonts}
\makeatletter
\@fleqnfalse
\@mathmargin\@centering
\makeatother

\emergencystretch=\maxdimen
\begin{document}
\title{Multi-scale Image Decomposition\\using a Local Statistical Edge Model}

\author{\IEEEauthorblockN{Kin-Ming Wong}
	\IEEEauthorblockA{\textit{School of Creative Media} \\
		\textit{City University of Hong Kong}\\
		Hong Kong S.A.R., China \\
		smmike@cityu.edu.hk}
}
\maketitle

\begin{abstract}
We present a progressive image decomposition method based on a novel non-linear filter named Sub-window Variance filter.  Our method is specifically designed for image detail enhancement purpose; this application requires extraction of image details which are small in terms of both spatial and variation scales.  We propose a local statistical edge model which develops its edge awareness using spatially defined image statistics.  Our decomposition method is controlled by \emph{two intuitive parameters} which allow the users to define what image details to suppress or enhance.  By using the summed-area table acceleration method, our decomposition pipeline is highly parallel. The proposed filter is gradient preserving and this allows our enhancement results free from the gradient-reversal artefact.  In our evaluations, we compare our method in various multi-scale image detail manipulation applications with other mainstream solutions.
\end{abstract}

\begin{IEEEkeywords}
	Multi-scale image decomposition, Image processing, Edge-preserving filter
\end{IEEEkeywords}

\section{Introduction}

Many image manipulation applications use the multi-scale image decomposition method \cite{farbman2008edge} for image detail enhancement. This kind of image decomposition is often represented by the following equation:
\begin{equation}
I = B + \sum_{i=1}^{N}D_i
\label{eqn_decomp_Intro}
\end{equation}
where $I$ is the original image. The decomposition process produces a base layer $B$ and multiple detail layers $D_i$.  The base layer is often a piecewise smooth image which maintains the low-frequency information and the primary image structures.  The detail layers $D_i$ hold image details of various scales.

\begin{figure}[H]
	\includegraphics[width=1.0\linewidth]{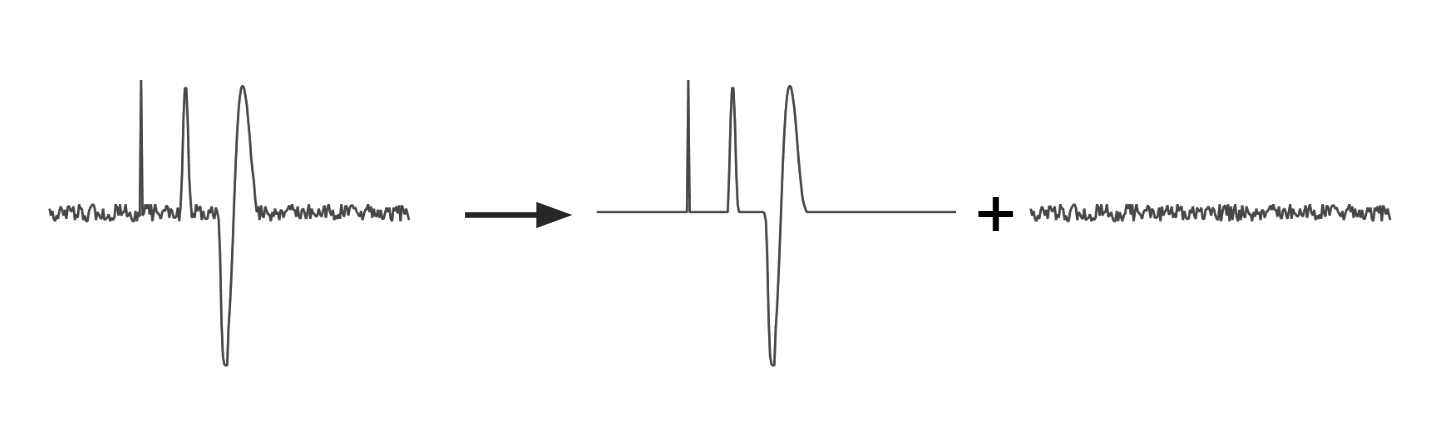}
	\caption{Fine details (in terms of both spatial \emph{and} signal variation scales) extraction from the input signal.}
	\label{fig:decomp}	
\end{figure}

Multi-scale image decomposition relies on coarsening an image with a smoothing filter to obtain a detail layer which is the difference between the original and the filtered image.  This process repeats with the filter support increased at each iteration, and stops when the desired number of detail layers have been obtained.  Fig. \ref{fig:decomp} shows an example of the decomposition process where the fine details which are small in both spatial and signal variation scales are separated.  

The early form of multi-scale image manipulation technique is based on Laplacian pyramid \cite{burt1983laplacian}. However, this non edge-preserving approach often produces halo artefacts, so edge-awareness becomes an essential requirement of image decomposition. Anisotropic diffusion \cite{perona1990scale} is a non-linear filtering method which uses the heat transfer model to detect edges, and there is an improved extension \cite{black1998robust} which includes the use of robust statistics.

The bilateral filter \cite{aurich1995non,tomasi1998bilateral} has become the mainstream edge-preserving filtering method. Many image decomposition and dynamic range compression related works \cite{bae2006two,Durand:2002:FBF:566570.566574,fattal2007multiscale} rely on the bilateral filter.  However, its over-sharpening characteristic is undesirable for certain image manipulation purposes, and is discussed in some recent studies \cite{farbman2008edge,he2013guided}.  Two essential properties of an edge-preserving filter suitable for multi-scale image decomposition application are identified by the Weighted Least Square (WLS) method article \cite{farbman2008edge}.  The authors emphasize that the decomposition filter should preserve important edge signals and it must be gradient preserving in order to avoid the over-sharpening effect.

In this paper, we propose a high quality edge-preserving local filter based on a novel local statistical edge model.  Our filter preserves user-defined edges practically unchanged, and it is gradient preserving.  We discuss a quality factor that has not been studied in previous literatures; it is \emph{the uniformity of spatial scale and signal variations in the detail layers}.  Our proposed method allows a flexible progressive decomposition without a constraint of the spatial scale increment as in wavelet based or other pyramidal schemes.

\section{Related Work}

The bilateral filter has many enhanced variants available  \cite{Chen:2007:REI:1275808.1276506,cho2014bilateral,Durand:2002:FBF:566570.566574,fattal2007multiscale,gavaskar2019fast,paris2006fast,xu2018improved,yang2009real}. However, most of them maintain the over-sharpening characteristic which often results gradient reversal artefacts in the detail enhanced images.  Many modern edge-preserving filters \cite{chang2015propagated,gastal2011domain,lee2017structure,zhang2014rolling} share this characteristic.  A thorough analysis of artefacts caused by using the bilateral filter and its variations is available \cite{farbman2008edge}.  The guided image filter \cite{he2013guided} has attractive performance but the authors recognize that its edge-preserving capability cannot eliminate the halo artefacts.

Optimization based filters rely on carefully designed constraints to achieve edge-preserving filtering using the global optimization approach. The weighted least square (WLS) framework \cite{farbman2008edge} overcomes the two artefacts which hinder most local filters.  However, we notice that the WLS method cannot differentiate spatially small-scale image structures from the less important image details which are small in both spatial and signal variation scales. Other optimization based filters \cite{gu2013local,xu2011image,xu2012structure} tend to summarize the piecewise smooth areas into step edges, i.e. non gradient-preserving.  In some cases, spatial displacement of features are observed \cite{hua2014edge}, and this side effect is not desirable for the image detail enhancement purpose.  Another practical concern about the optimization based filters is the opaqueness of the filter parameters. Unlike the ones in kernel based filters, the parameters in the optimization based filters are often non-intuitive from an end-user's perspective.

Wavelet based solution \cite{fattal2009edge} offers good edge-aware decomposition but the rigid decimation of spatial scale prevents flexible manipulation of image details.  The more recent deep learning approach \cite{zhu2019benchmark} learns the filtering results from a selection of curated filtered images. As a result, their effectiveness for image detail enhancement becomes largely dependent on the training data set.
\\
\section{Sub-window Variance Filter}

\subsection{Local Statistical Edge Model}
We propose an edge model which uses simple statistical information to identify edge-alike features on local image patches.  Our model comprises three conditions as follows:
\begin{enumerate}
	\item An edge is formed by two adjacent groups of pixels.
	\item These two groups have contrasting intensities (global high variance).
	\item An edge is more distinct if the pixels of either group share similar intensity (locally low variance in a subregion).
\end{enumerate}

\begin{figure}	
	\begin{center}		
		\begin{subfigure}[t]{0.46\linewidth}
			\centering
			\includegraphics[width=1.0\linewidth]{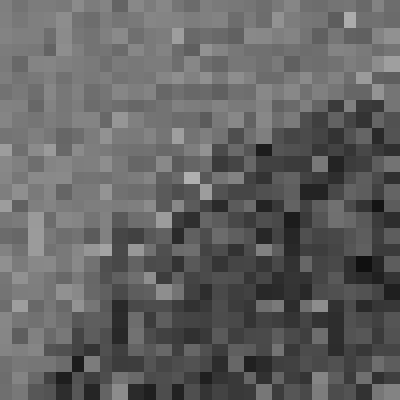}
			\caption{A fuzzy edge formed by two contrasting groups of pixels exhibiting overall high variance}
			\label{fig_EdgeFuzzy}			
		\end{subfigure}
		\hspace{0.2cm}
		\begin{subfigure}[t]{0.46\linewidth}
			\centering
			\includegraphics[width=1.0\linewidth]{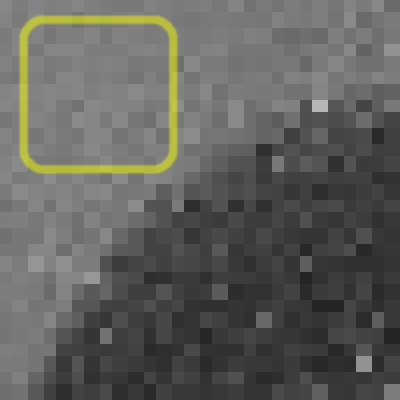}
			\caption{A distinct edge formed by two contrasting groups with one group exhibits low variance (highlighted)}
			\label{fig_EdgeDistinct}
		\end{subfigure}
		\caption{Presence of a relatively low variance sub-region as an evidence of the presence of edge(s)}
	\end{center}
	\label{f2_EdgeModel}
\end{figure}

The first condition states the importance of the spatial property required to define an edge, and the second condition describes the signal strength difference necessary for an edge to be perceived.  Given an image patch which contains a distinct edge, the global statistical variance of intensity values should be high as suggested by the second condition.  However, a patch which has a high variance value does not necessarily imply the existence of an edge due to the lack of spatial information.  Fig. \ref{fig_EdgeFuzzy} illustrates an image patch containing a fuzzy edge which exhibits high variance.

The most important observation we made about the presence of a distinct edge is stated in the third condition.  Fig. \ref{fig_EdgeDistinct} illustrates such a distinct edge; it is the presence of a subregion which has a low local variance serves as a clear background to highlight the adjacency of two groups of pixels.  In order to validate our edge model, we present a simple analysis in next section to show how the statistical variance values of the local sub-regions, i.e. how the sub-window variance values relate to the global variance.

\subsection{Properties of Sub-window Variance}
\begin{figure}
	\begin{center}		
		\includegraphics[width=0.9\linewidth]{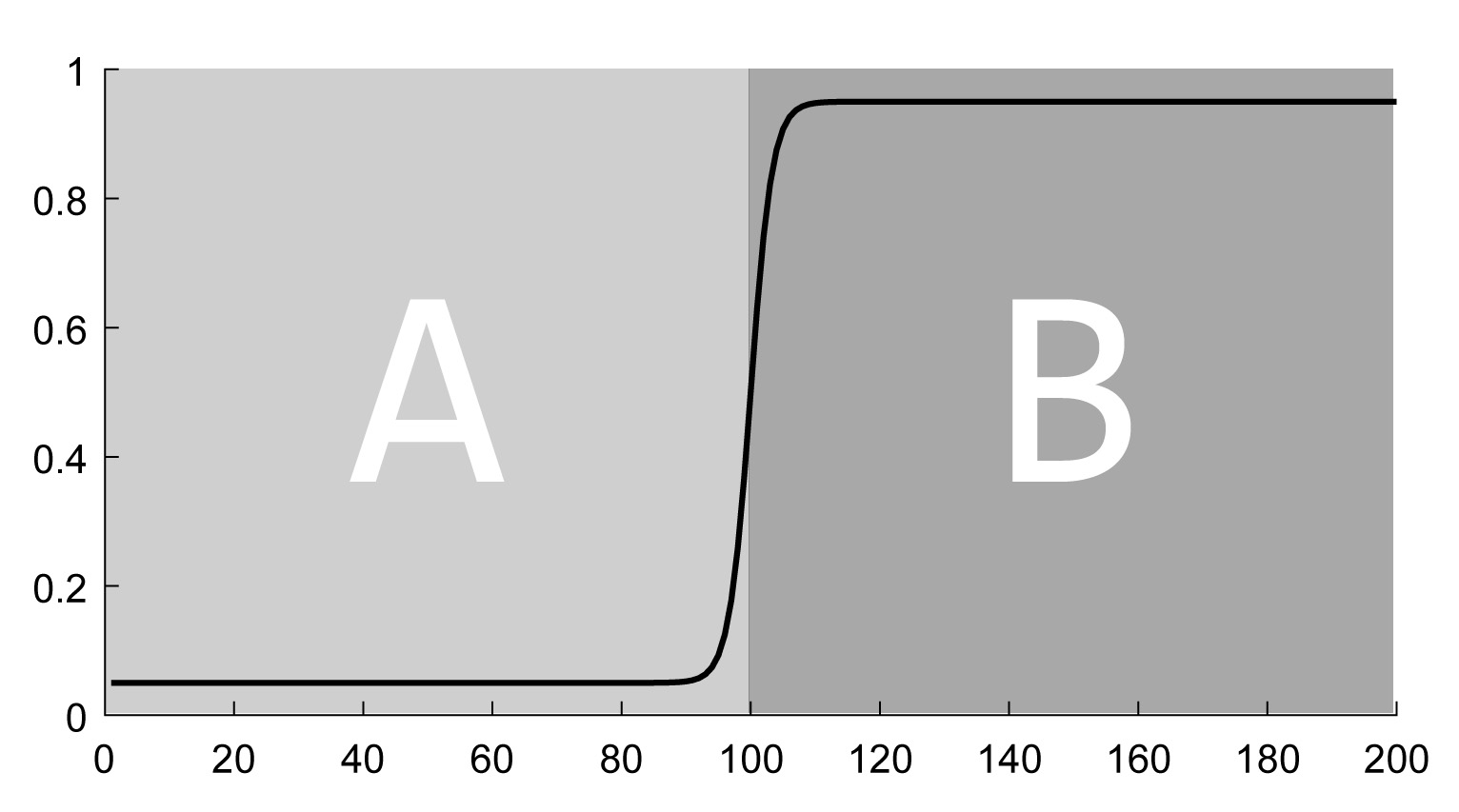}
		\caption{A simple 1D signal and the two equal-sized sub-windows A and B}
		\label{fig_1D_signal}			
	\end{center}
\end{figure}

By definition, the variance of a discrete signal is the expected value of the squared deviation from the mean value of the signal.  The variance value of any arbitrary rectangular image patch on an image can be computed quickly by using a pair of precomputed summed-area tables \cite{burger2016digital,Crow:1984:STT:800031.808600}.
Let us consider a simple 1D signal $X$ as shown in Fig. \ref{fig_1D_signal}, and we use two equal-sized sub-windows to divide the signal into two halves namely $X_A$ and $X_B$.  We express the expected value of $X$ as the mean of the expected values evaluated from the two sub-windows as expressed in equation (\ref{eqn_sum_subWinMean}).
\begin{equation}
\label{eqn_sum_subWinMean}
E[X] = \frac{E[X_A]}{2} + \frac{E[X_B]}{2}
\end{equation}
By using equation (\ref{eqn_sum_subWinMean}), we explore the relationship between the overall variance $Var(X)$ and the sub-window variance values $Var(X_A)$ and $Var(X_B)$ through the steps as shown in equation (\ref{eqn_SubWinVar_Definition}).
\\
\begin{align}
\begin{split}
\label{eqn_SubWinVar_Definition}
Var(X)&= E[X^2] - (E[X])^2\\
&= \frac{E[X_A^2]}{2} + \frac{E[X_B^2]}{2} - \Big(\frac{E[X_A]}{2} + \frac{E[X_B]}{2}\Big)^2\\
&= \frac{E[X_A^2]-(E[X_A])^2}{2} + \frac{E[X_B^2]-(E[X_B])^2}{2}\\
&~~~+\Big(\frac{E[X_A]}{2} - \frac{E[X_B]}{2}\Big)^2\\
&= \frac{Var(X_A)}{2} + \frac{Var(X_B)}{2} + \frac{(E[X_A]-E[X_B])^2}{4}
\end{split}
\end{align}

For a given variance value $Var(X)$ of a signal, if both sub-window variance values $Var(X_A)$ and $Var(X_B)$ are small or close to zero, then according to equation (\ref{eqn_SubWinVar_Definition}), the difference between $E[X_A]$ and $E[X_B]$ is maximized and therefore it concludes the presence of a clear step edge feature.  This important property allows us to use sub-window variance values as a means to detect edge-alike features in a signal.  

Furthermore, according to equation (\ref{eqn_SubWinVar_Definition}), when the values of $Var(X)$, $Var(X_A)$ and $Var(X_B)$ are identical, the third term vanishes.  For the third term to become zero, $E[X_A]$ must be equal to $E[X_B]$ and by equation (\ref{eqn_sum_subWinMean}), it further concludes that all three expected values $E[X]$, $E[X_A]$ and $E[X_B]$ must be equal too.  One possible interpretation is the presence of self-similarity such as a constant flat signal or the presence of regular patterns with its feature wavelength smaller than the filter support.

\subsection{Sub-window Variance Filter}
We formulate the edge-preserving filtering process on an image patch as a linear blend of the original patch and a filtered version of the patch.  For a given local patch $I_k$ centred on pixel $k$, our filtered patch $I_k^\prime$ is computed by:
\begin{equation}
I_k^\prime=A_kI_k + (1-A_k)F(I_k),	
\label{eqn_PrimeFilterModel}
\end{equation}
where $A_k\in[0,1]$. $A_k$ is called the \emph{per-patch preservation factor} which governs the degree of contribution by the original patch. $F()$ is a generic smoothing filter, and we choose the box filter for its simplicity which evaluates the mean intensity values of the original patch.  This simple linear blend formulation forces the filtered value bound to the original value and the mean of the whole patch.  We rely on this formulation to eliminate the possibility of over-sharpening.  We rewrite equation (\ref{eqn_PrimeFilterModel}) as follows:
\begin{equation}
I_k^\prime=A_kI_k + B_k,	
\label{eqn_SimplifiedFilterModel}
\end{equation}
where $B_k=(1-A_k)(\frac{1}{|\omega|}\sum\limits_{i\in{\omega_k}}p_i)$, $\omega_k$ is the local domain defined by the filter support centred at pixel $k$, and $p_i$ is the intensity of the $i^{th}$ pixel.

\begin{figure}
	\begin{center}		
		\begin{subfigure}[t]{0.325\linewidth}
			\centering
			\includegraphics[width=1.0\linewidth]{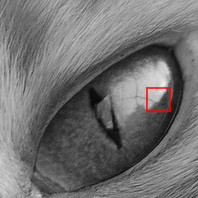}
			\caption{Local image patch highlighted}
			\label{fig_CatEye}			
		\end{subfigure}
		\hfill
		\begin{subfigure}[t]{0.325\linewidth}
			\centering
			\includegraphics[width=1.0\linewidth]{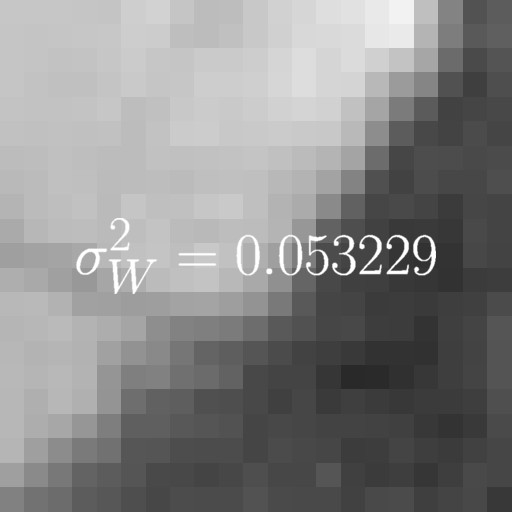}
			\caption{image patch variance value}
			\label{fig_CatEye_varW}			
		\end{subfigure}
		\hfill
		\begin{subfigure}[t]{0.325\linewidth}
			\centering
			\includegraphics[width=1.0\linewidth]{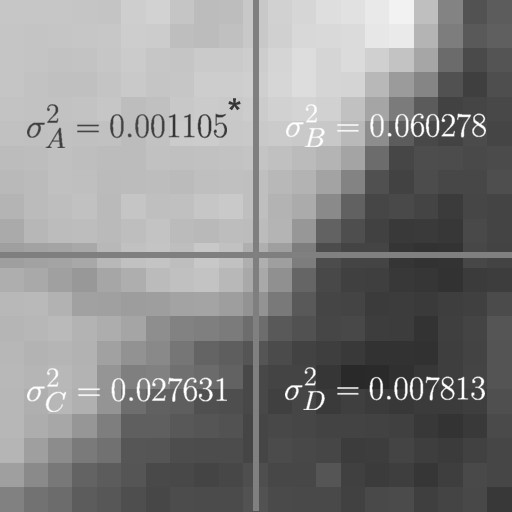}
			\caption{Sub-window variance values}
			\label{fig_CatEye_varABCD}
		\end{subfigure}
		\caption{Edge model exploiting locally low variance subregion as an evidence of presence of an edge}
	\end{center}
	\vspace{0.0cm}
\end{figure}

\textbf{Per-patch Preservation Factor}.  Based on our proposed edge model, we divide the filter window into four equal-sized sub-windows namely A, B, C and D (Fig. \ref{fig_CatEye_varABCD}) in order to evaluate the variance value of each sub-window.  This simple uniform sampling strategy is motivated by the advantage of using summed-area table \cite{burger2016digital,Crow:1984:STT:800031.808600} for high performance filtering. This subdivision scheme also resembles the one used in the Kuwahara filter \cite{kuwahara1976processing}.

Let $V=\{\sigma_A^2,\sigma_B^2,\sigma_C^2,\sigma_D^2\}$ be the set of intensity variances evaluated on the four sub-windows, and $\sigma_W^2$ be the intensity variance of the whole patch.  We define the value of \emph{per-patch preservation factor}, $A_k$ by:	
\begin{equation}
A_k = min( 1, \frac{\sigma_{max}^2}{\sigma^2_{min}+\epsilon}),
\label{eqn_Ak}
\end{equation}
where $\sigma_{max}^2 = max(\{\sigma_W^2,V\})$, $\sigma_{min}^2 = min(V)$ and $\epsilon$ is a positive valued user parameter.

Based on the second condition of our edge model (the requirement of contrasting intensities), we use $\sigma_{max}^2$ as a potential edge indicator on the patch.  At the same time, we include the minimum sub-window variance value $\sigma_{min}^2$ in the denominator of equation (\ref{eqn_Ak}) to promote strong preservation of edges following the third condition of our edge model.

The user parameter $\epsilon$ has an intuitive meaning in this formulation.  Suppose we have a distinct edge with $\sigma_{min}^2 = 0$ and if $\sigma_{max}^2$ is equal to the value of $\epsilon$, then the whole patch will be fully preserved with $A_k = 1$.  In short, $\epsilon$ defines the threshold variance value of a clear edge in a given filter window.  In addition, our simple filter model guarantees two neighbour points on a gradient shall only have their intensity values become closer to each other when filtered, i.e. there is no risk of over-sharpening.

\begin{figure}
	\begin{center}		
		\begin{subfigure}[t]{0.325\linewidth}
			\centering
			\includegraphics[width=1.0\linewidth]{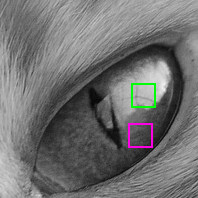}
			\caption{Local image patches highlighted}
			\label{fig_CatEye_LF_Patch}			
		\end{subfigure}
		\hfill
		\begin{subfigure}[t]{0.325\linewidth}
			\centering
			\includegraphics[width=1.0\linewidth]{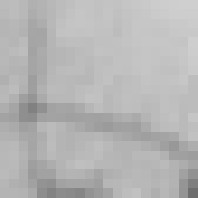}
			\caption{Thin lines \\$\sigma_W^2 = 0.003327$}
			\label{fig_CatEye_Line}			
		\end{subfigure}
		\hfill
		\begin{subfigure}[t]{0.325\linewidth}
			\centering
			\includegraphics[width=1.0\linewidth]{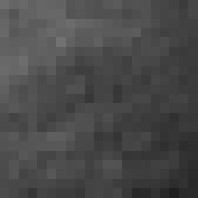}
			\caption{Fuzzy\\$\sigma_W^2 = 0.003342$}
			\label{fig_CatEye_Fuzzy}
		\end{subfigure}
		\caption{Feature sensitivity of sub-window variance filter}
	\end{center}
	\vspace{0.0cm}
	\label{fig_CatEye_LF}
\end{figure}

Although our theoretical development focuses on edges, the formulation is not biased on any geometrical property. We extract two patches from Fig. \ref{fig_CatEye_LF_Patch}, and while they look seemingly different (Fig. \ref{fig_CatEye_Line} and \ref{fig_CatEye_Fuzzy}), they have numerically close variance values.  Sub-window variance values and per-patch preservation factors of both patches are computed and tabulated in Tables \ref{tab_varABCD} and \ref{tab_Ak}.  The computed per-patch preservation factors successfully differentiate these two patches, and conclude that the thin lines patch deserves stronger preservation.

\newcolumntype{Z}{>{\centering\arraybackslash\hspace{0pt}}X}
\begin{table}[H]
	\centering
	\caption{Sub-window variance values of patches in Fig. \ref{fig_CatEye_Line} (Thin lines) and \ref{fig_CatEye_Fuzzy} (Fuzzy)}
	\vspace{-0.1cm}
	\label{TabSubWinVar}
	\def\arraystretch{1.5}
	\begin{tabular}{ |M{1.2cm}|M{1.25cm}|M{1.25cm}|M{1.25cm}|M{1.25cm}|N }
		\hline
		Patch & ${\sigma_A^2}$ & ${\sigma_B^2}$ & ${\sigma_C^2}$ & ${\sigma_D^2}$ &\\
		\hline
		Lines & 0.001352 & $0.000375^*$ & 0.003283 & 0.003011 &\\
		\hline
		Fuzzy & 0.002491 & 0.001343 & $0.001061^*$ & 0.001369 &\\
		\hline
	\end{tabular}
	\label{tab_varABCD}
\end{table}

\begin{table}[H]
	\centering
	\caption{Per-patch preservation using equation (\ref{eqn_Ak}), $\epsilon = 0.0028$ }
	\vspace{-0.1cm}
	\label{TabPerPatch}
	\def\arraystretch{1.5}
	\begin{tabular}{ |M{1.2cm}|M{1.2cm}|M{1.2cm}|M{2.7cm}|N }
		\hline
		Patch  & $\sigma_{min}^2$ & $\sigma_{max}^2$ & $A_k$ & \\
		\hline
		Lines  & 0.000375 & 0.003327 & $\min(1, 1.0479)$ & \\
		\hline
		Fuzzy  & 0.001061 & 0.003342 & $\min(1, 0.8656)$ & \\
		\hline	
	\end{tabular}
	\label{tab_Ak}
\end{table}

\begin{figure}
	\centering
	\includegraphics[trim={0 1.5cm 0 2cm},clip,width=1.0\linewidth]{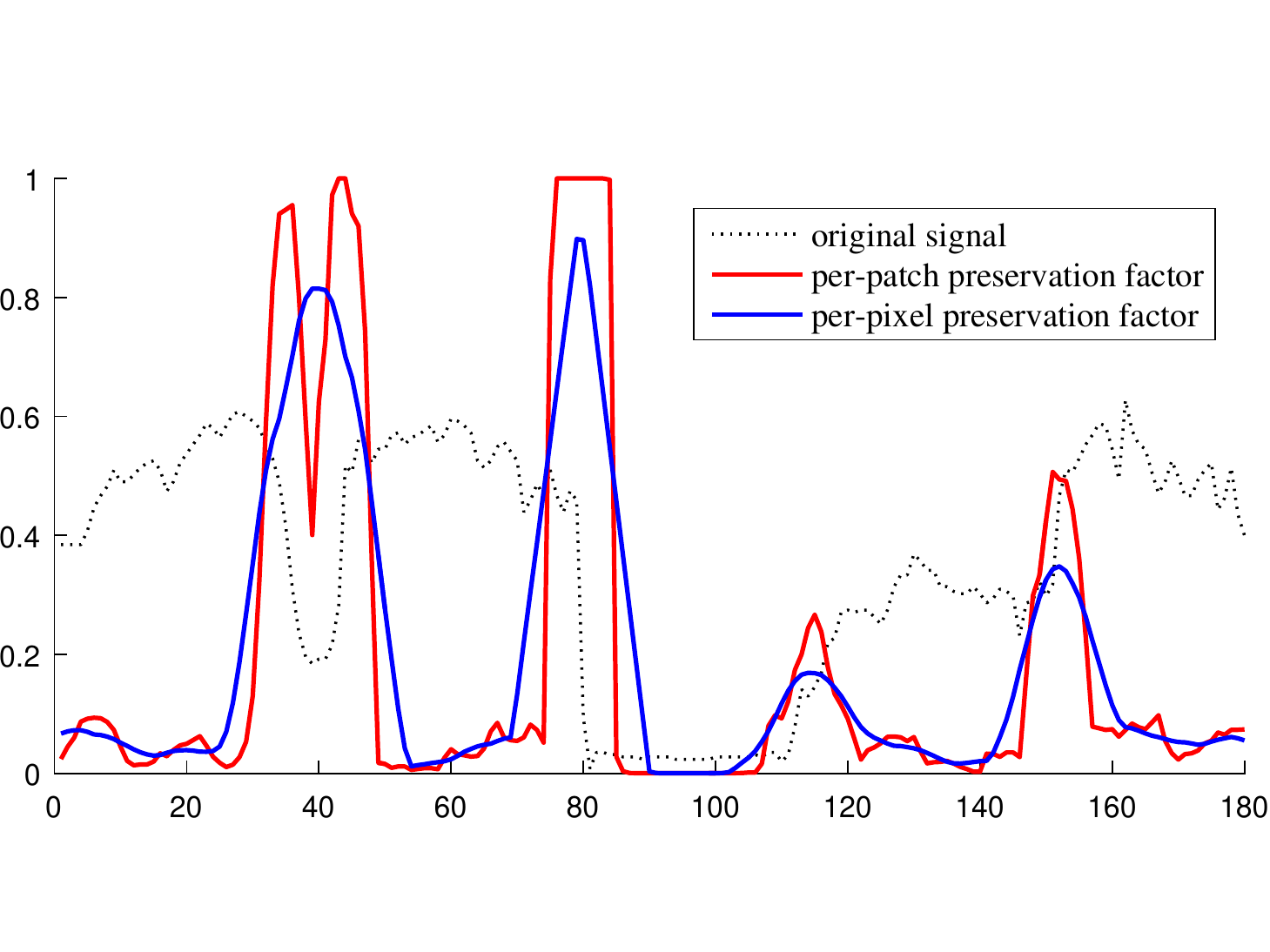}
	\caption{Per-patch and per-pixel preservation factors\\ $filter~width = 11,~\epsilon = 0.025$}
	\label{fig_preserveFactors}			
\end{figure}

\textbf{Per-pixel preservation factor for image filtering.} In order to filter a whole image, we densely evaluate the per-patch preservation factors over the whole image.  Since each pixel receives multiple per-patch preservation estimations from a fixed number of overlapping patches, the final filtered output is given by:
\begin{equation}
p_i^\prime=\bar{A_k}p_i + \bar{B_k},	
\label{eqn_pixelFilter}
\end{equation}
where $\bar{A_k}=\frac{1}{|\omega|}\sum\limits_{k\in{\omega_i}}A_k$, $\bar{B_k}=\frac{1}{|\omega|}\sum\limits_{k\in{\omega_i}}B_k$ and $\omega_i$ is the domain of all overlapping patches that offer estimation to pixel $i$. $\bar{A_k}$ can be understood as a \emph{per-pixel preservation factor}.  This simple averaging strategy guarantees piece-wise continuity, and is a common practice among modern image filters \cite{dabov2007image,gu2013local,he2013guided,katkovnik2010local}.  Fig. \ref{fig_preserveFactors} illustrates how the per-pixel preservation factors (blue line) accurately capture the strong edge locations of a given 1D signal.
\\

\begin{figure*}
	\begin{center}	
		\begin{subfigure}[t]{0.192\linewidth}
			\centering
			\includegraphics[width=1.0\linewidth]{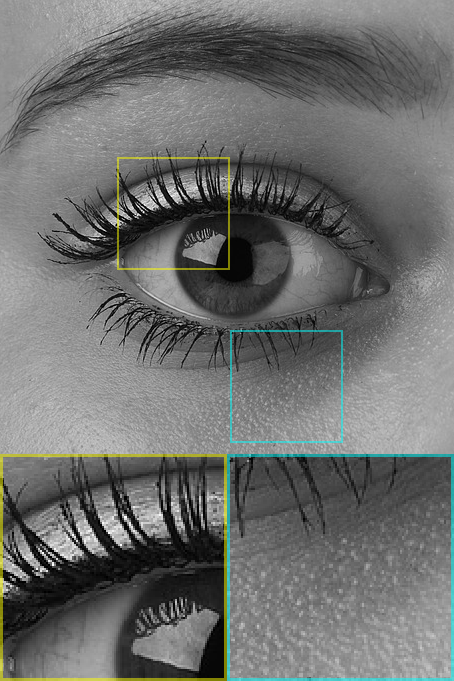}
			\caption{Original}
			\label{fig_eye3_org}			
		\end{subfigure}
		\begin{subfigure}[t]{0.192\linewidth}
			\centering
			\includegraphics[width=1.0\linewidth]{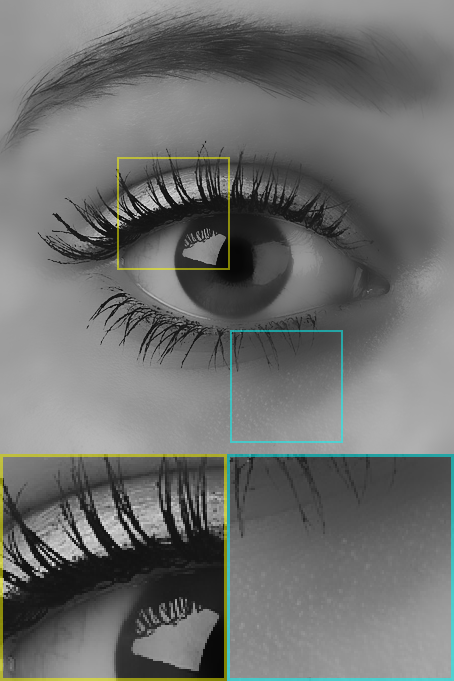}
			\caption{Bilateral filter\\$\sigma_d=10,~\sigma_r=0.1$\\SSIM = 0.76319}			
			\label{fig_eye3_BF}			
		\end{subfigure}
		\begin{subfigure}[t]{0.192\linewidth}
			\centering
			\includegraphics[width=1.0\linewidth]{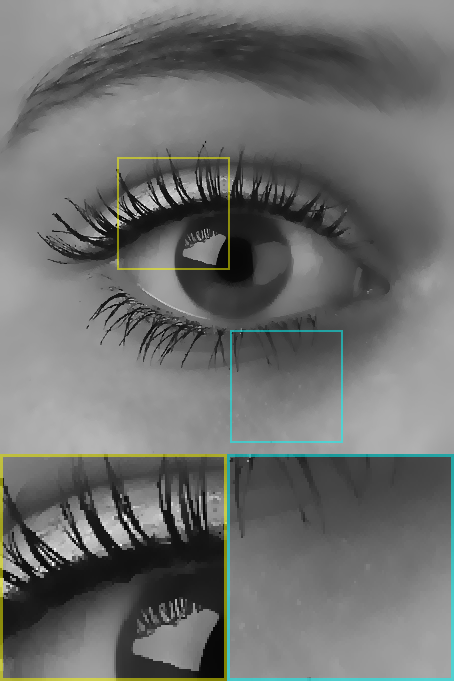}
			\caption{Domain Transform\\$\sigma_s=10,~\sigma_r=0.45^*$\\SSIM = 0.64674}
			\label{fig_eye3_DT}
		\end{subfigure}
		\begin{subfigure}[t]{0.192\linewidth}
			\centering
			\includegraphics[width=1.0\linewidth]{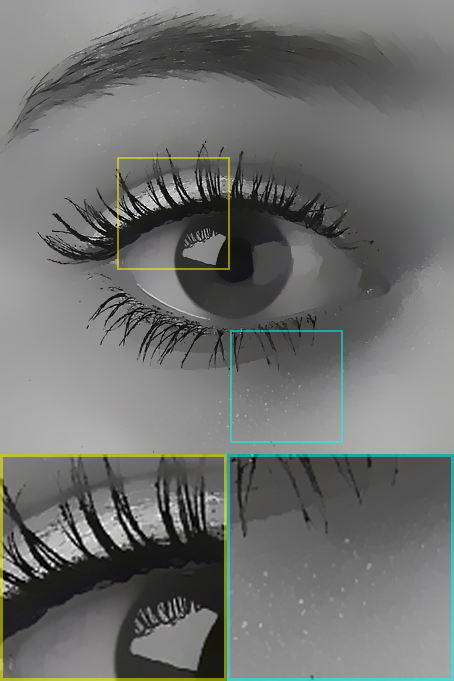}
			\caption{ResNet\\SSIM = 0.69766}
			\label{fig_eye3_ResNet}
		\end{subfigure}
		\begin{subfigure}[t]{0.192\linewidth}
			\centering
			\includegraphics[width=1.0\linewidth]{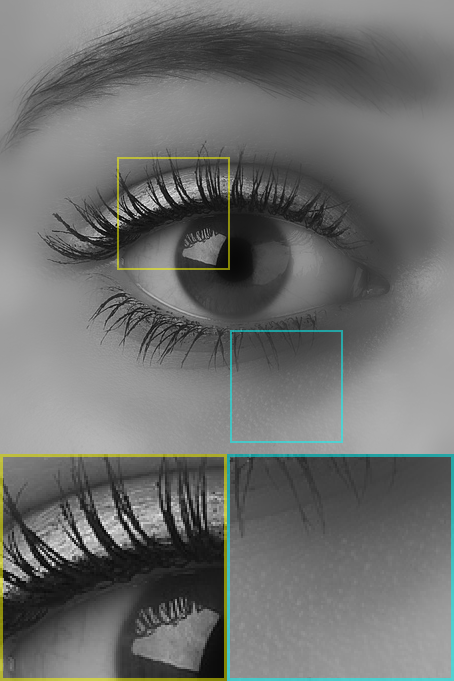}
			\caption{Guided filter\\$radius=10, ~\epsilon=0.01$\\SSIM = 0.76527}
			\label{fig_eye3_GIF}
		\end{subfigure}
		\vspace{0.1cm}	
		
		\hspace{0.192\linewidth}
		\begin{subfigure}[t]{0.192\linewidth}
			\centering
			\includegraphics[width=1.0\linewidth]{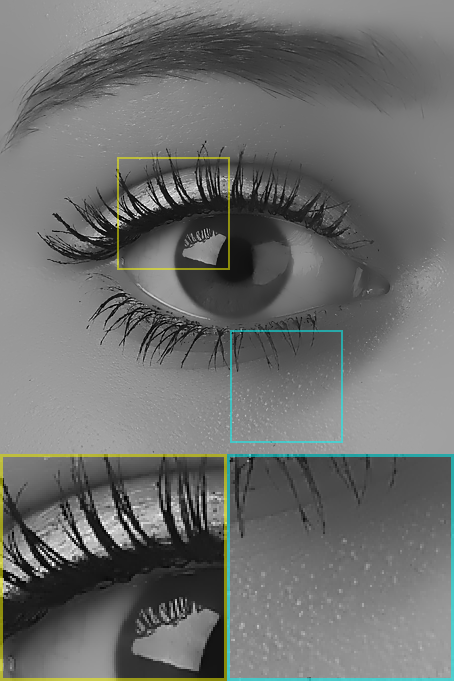}
			\caption{Local Laplacian\\$\sigma_r=0.28, \alpha=3$\\SSIM = 0.81236}
			\label{fig_eye3_FL}
		\end{subfigure}	
		\begin{subfigure}[t]{0.192\linewidth}
			\centering
			\includegraphics[width=1.0\linewidth]{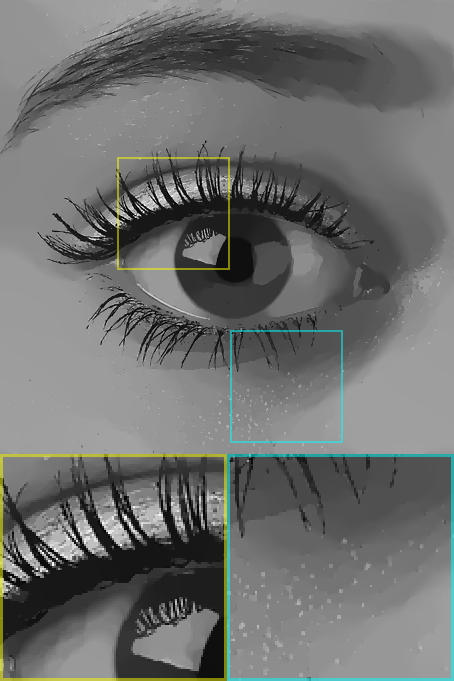}
			\caption{$L_0$ smoothing\\$\lambda=0.01$\\SSIM = 0.71559}
			\label{fig_eye3_L0S}			
		\end{subfigure}
		\begin{subfigure}[t]{0.192\linewidth}
			\centering
			\includegraphics[width=1.0\linewidth]{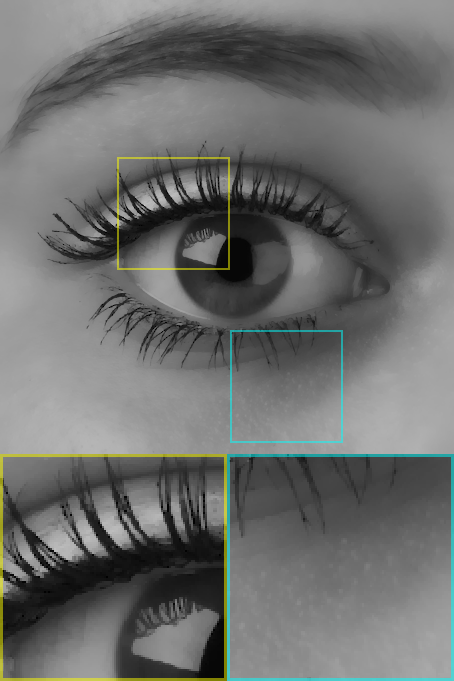}
			\caption{WLS\\$~\alpha=0.9, \lambda=0.6$\\SSIM = 0.70515}
			\label{fig_eye3_WLS}
		\end{subfigure}
		\begin{subfigure}[t]{0.192\linewidth}
			\centering
			\includegraphics[width=1.0\linewidth]{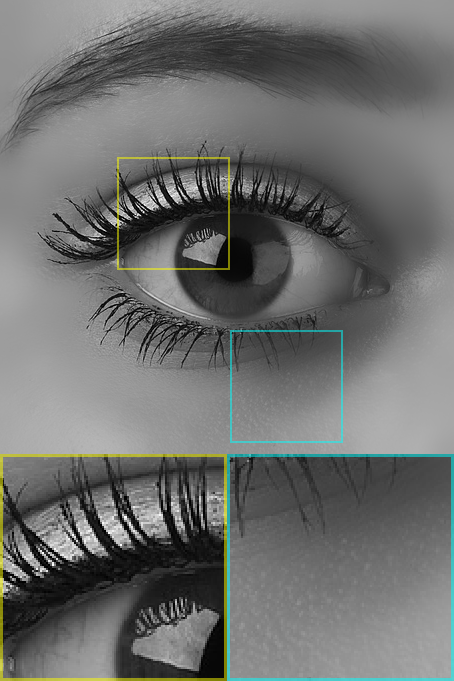}
			\caption{Ours\\$radius=10, ~\epsilon=0.01$\\SSIM = 0.82533}
			\label{fig_eye3_SVF}
		\end{subfigure}		
		\caption{Edge-preserving and gradient-preserving performance comparison.  All filter parameters are set according to the approach of matching the bilateral filter documented in the respective literatures; *Parameters of the Domain Transform filter were hand tuned for improved smoothing performance.}
		\label{fig_eye3}
	\end{center}
	\vspace{0.0cm}
\end{figure*}

\subsection{Sub-window variance filter characteristics}
\begin{figure}
	\begin{center}	
		\begin{subfigure}[t]{0.32\linewidth}
			\centering
			\includegraphics[width=1.0\linewidth]{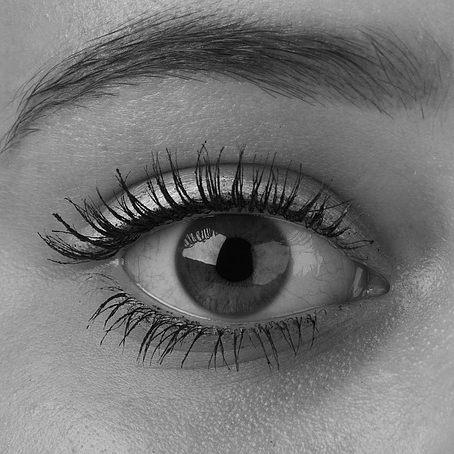}
			\caption{Original}			
			\label{fig_diff_ORG}			
		\end{subfigure}
		\begin{subfigure}[t]{0.32\linewidth}
			\centering
			\includegraphics[width=1.0\linewidth]{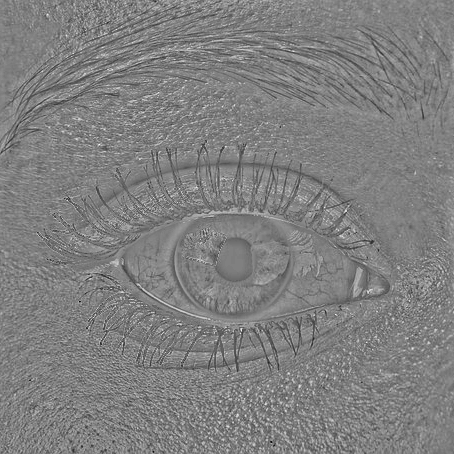}
			\caption{Bilateral filter}			
			\label{fig_diff_BF}			
		\end{subfigure}
		\begin{subfigure}[t]{0.32\linewidth}
			\centering
			\includegraphics[width=1.0\linewidth]{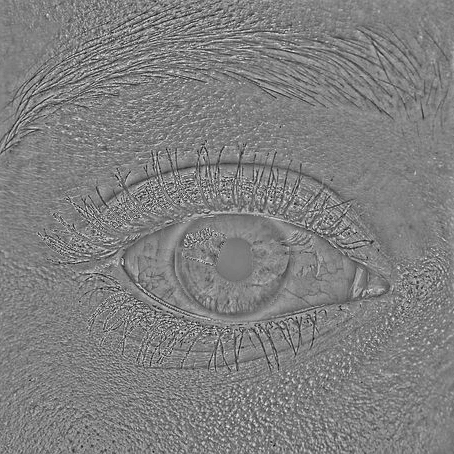}
			\caption{Domain transform}
			\label{fig_diff_DT}
		\end{subfigure}
		\begin{subfigure}[t]{0.32\linewidth}
			\centering
			\includegraphics[width=1.0\linewidth]{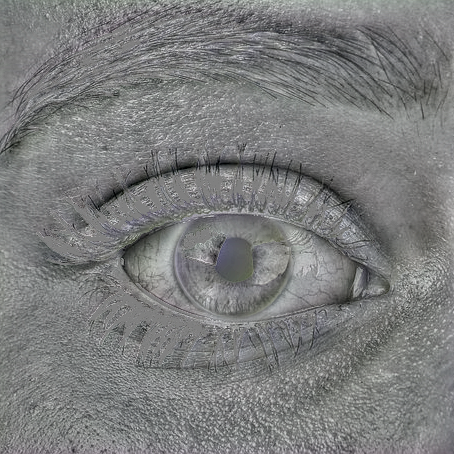}
			\caption{ResNet}
			\label{fig_diff_ResNet}
		\end{subfigure}
		\begin{subfigure}[t]{0.32\linewidth}
			\centering
			\includegraphics[width=1.0\linewidth]{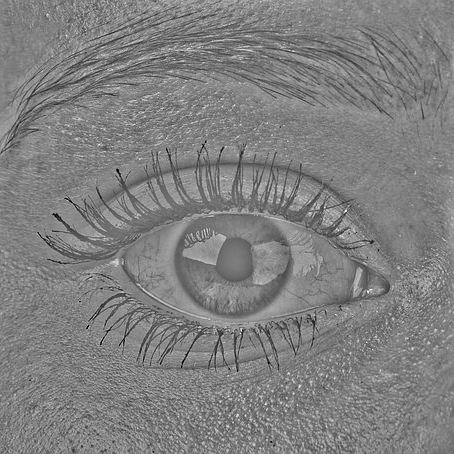}
			\caption{Guided filter}
			\label{fig_diff_GIF}
		\end{subfigure}
		\begin{subfigure}[t]{0.32\linewidth}
			\centering
			\includegraphics[width=1.0\linewidth]{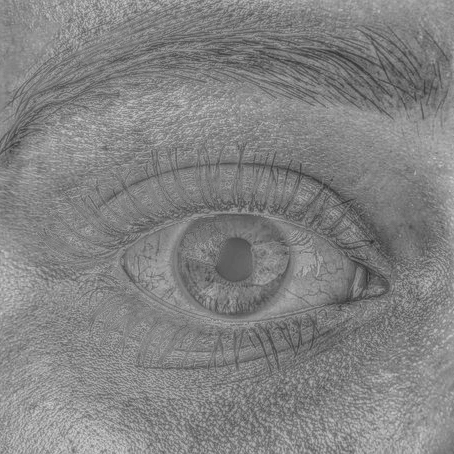}
			\caption{Local Laplacian}
			\label{fig_diff_LP}
		\end{subfigure}	
		\begin{subfigure}[t]{0.32\linewidth}
			\centering
			\includegraphics[width=1.0\linewidth]{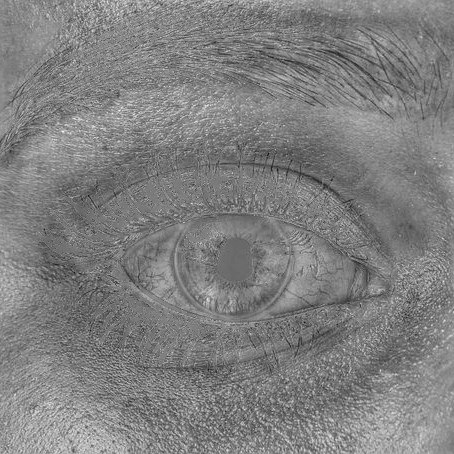}
			\caption{$L_0$ smoothing}
			\label{fig_diff_L0S}			
		\end{subfigure}
		\begin{subfigure}[t]{0.32\linewidth}
			\centering
			\includegraphics[width=1.0\linewidth]{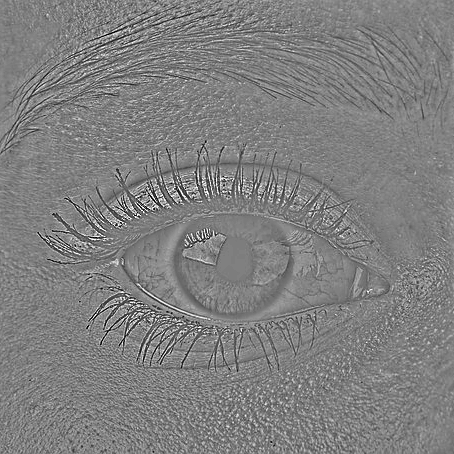}
			\caption{WLS}
			\label{fig_diff_WLS}
		\end{subfigure}
		\begin{subfigure}[t]{0.32\linewidth}
			\centering
			\includegraphics[width=1.0\linewidth]{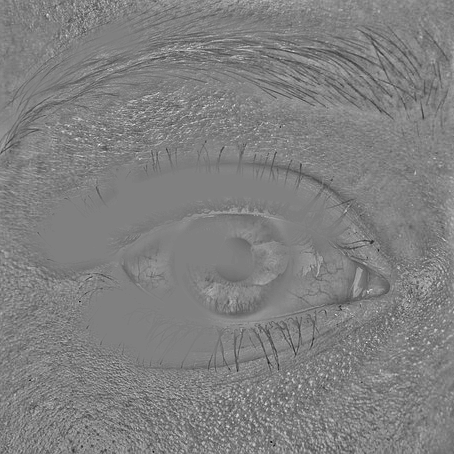}
			\caption{Ours}
			\label{fig_diff_SVF}
		\end{subfigure}
		\hspace{0.32\linewidth}	
		\caption{Detail layers extracted by various filters.  Our method (i) does not extract the eyelashes, and other important image structures into the detail layer.}
		\label{fig_eye3_diff}
	\end{center}
	\vspace{0.0cm}
\end{figure}

\textbf{Basic smoothing and detail layer properties.} We compare our filter with the leading edge-preserving filters commonly used in image decomposition (Fig.~\ref{fig_eye3}), such as the bilateral filter \cite{tomasi1998bilateral} and the WLS smoothing method \cite{farbman2008edge}.  We have included several modern edge-preserving filters in our comparison.  Fig.~\ref{fig_eye3} shows the smoothing results from each filter, and Fig.~\ref{fig_eye3_diff} shows their respective extracted detail layers.

\begin{figure}
	\centering
	\includegraphics[width=1.0\linewidth]{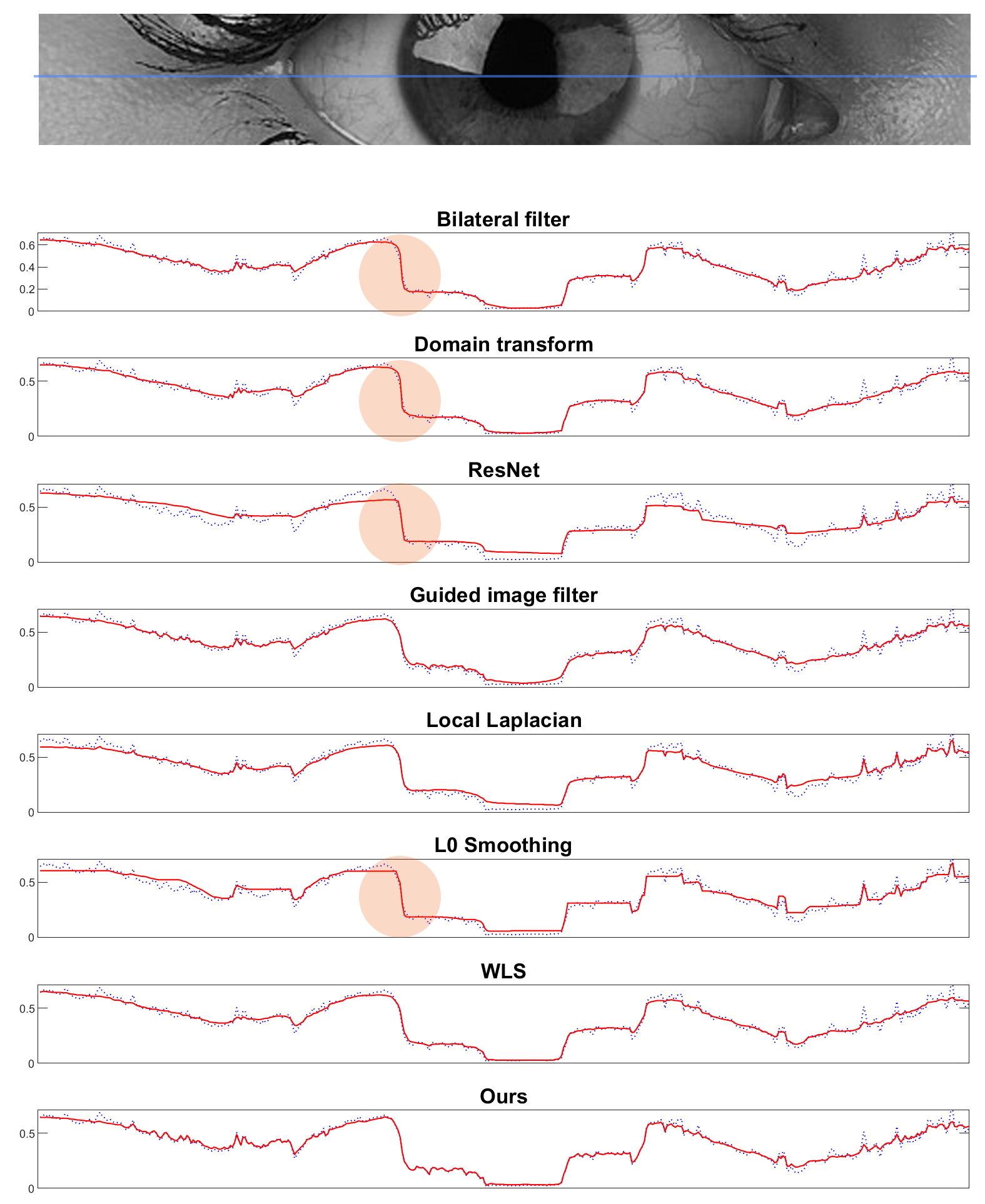}
	\caption{Close examination of the 1D slices extracted from the filtering results presented in Fig.\ref{fig_eye3}.  The blue dotted line represents the original signal, and the red ones represent the filtered signals.  The over-sharpened regions are highlighted with red coloured circles. }
	\label{fig_1D_summary}
\end{figure}

Fig.~\ref{fig_eye3_BF} shows the result using the bilateral filter, and it is the control that all filters attempt to match.  The domain transform (normalized convolution) filter \cite{gastal2011domain} (Fig.~\ref{fig_eye3_DT}) demonstrates satisfactory smoothing performance but it results over-sharpening on the edges.  Fig.~\ref{fig_diff_BF} and \ref{fig_diff_DT} show the detail layers extracted by these filters; they contain features are small in both spatial and variation scales.  However, they both exhibit over-sharpening characteristics on the edge of the pupil as shown in Fig.~\ref{fig_1D_summary}. 

The ResNet based filtering \cite{zhu2019benchmark} result shown in Fig.~\ref{fig_eye3_ResNet} exhibits characteristics shared by most optimization based filters.  There are step edges and loss of the important gradients especially on the eyelid.  Among all the detail layers produced, the one by ResNet (Fig.~\ref{fig_diff_ResNet}) exhibits a strong presence of image structures. The guided filter \cite{he2013guided} (Fig.~\ref{fig_eye3_GIF}) delivers reasonable smoothing but there is an image-wide contrast reduction due to weak edge preservation.  This will likely lead to halos when the detail layers are aggressively enhanced. Fig.~\ref{fig_diff_GIF} shows that its detail layer carries important image structures.  

Local Laplacian filtering \cite{aubry2014fast,paris2011local} has difficulty to deliver comparable smoothing without aggressive filtering.  The result (Fig.~\ref{fig_eye3_FL}) has considerable amount of medium-scale gradients removed.  Its detail layer (Fig.~\ref{fig_diff_LP}) also suggests the presence of strong image structures.  We have mentioned that most structure extraction filters \cite{xu2011image,xu2015deep,xu2012structure} are unsuitable for multi-scale image decomposition, and we evaluate the $L_0$ smoothing filter \cite{xu2011image} in order to illustrate our argument.  Fig.~\ref{fig_eye3_L0S} shows that the result is over-sharpened, and the banded edges are the natural consequence of its approximation based formulation.  Fig.~\ref{fig_diff_L0S} further shows the non-uniform detail layer which exhibits both image structures and low frequency information.

WLS filtering approach \cite{farbman2008edge} is proposed specifically for multi-scale image decomposition.  Its smoothing result (Fig.~\ref{fig_eye3_WLS}) matches the bilateral filter but there is an obvious contrast reduction on the strong edges.  Fig.~\ref{fig_diff_WLS} shows that its respective detail layer holds structural edges, and it will result considerable halos upon detail enhancement.

Fig.~\ref{fig_eye3_SVF} shows our filtering result, and it matches closely the smoothing result of the bilateral filter. Our result is free from any over-sharpening or contrast reduction.  Fig.~\ref{fig_diff_SVF} further illustrates the strong edge-preservation capability where the high-contrast area of the iris and the eyelashes are practically untouched, and the detail layer demonstrates uniform fine scale details.

This basic comparison helps us to identify the candidates for more in-depth comparisons in the context of multi-scale image decomposition.  We shall now focus on the bilateral filter and the multi-scale specific WLS approach for further comparisons.

\begin{figure*}[t]
	\begin{center}	
		\begin{subfigure}[t]{0.325\linewidth}
			\centering
			\includegraphics[width=1.0\linewidth]{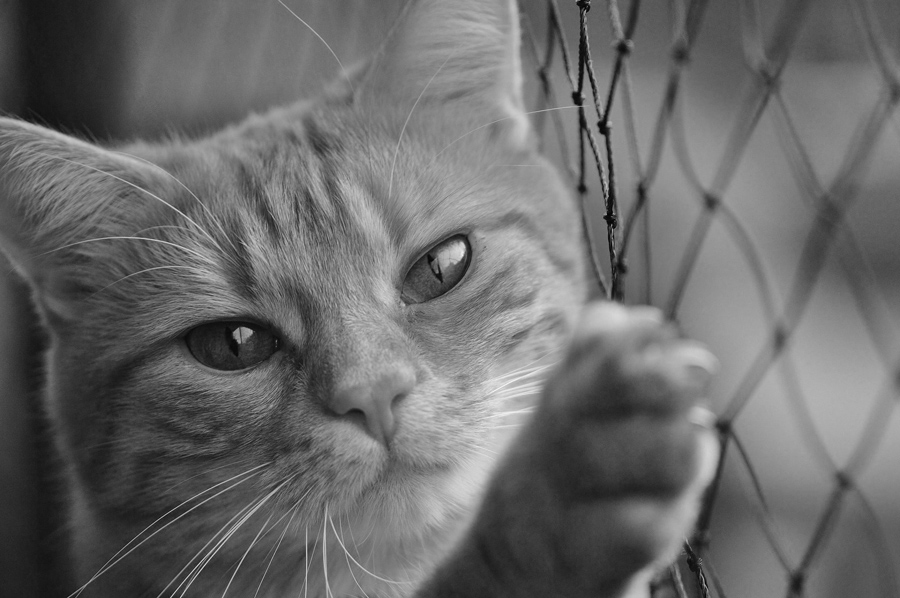}
			\caption{Original}
			\label{fig_cat_org}			
		\end{subfigure}
		\begin{subfigure}[t]{0.325\linewidth}
			\centering
			\includegraphics[width=1.0\linewidth]{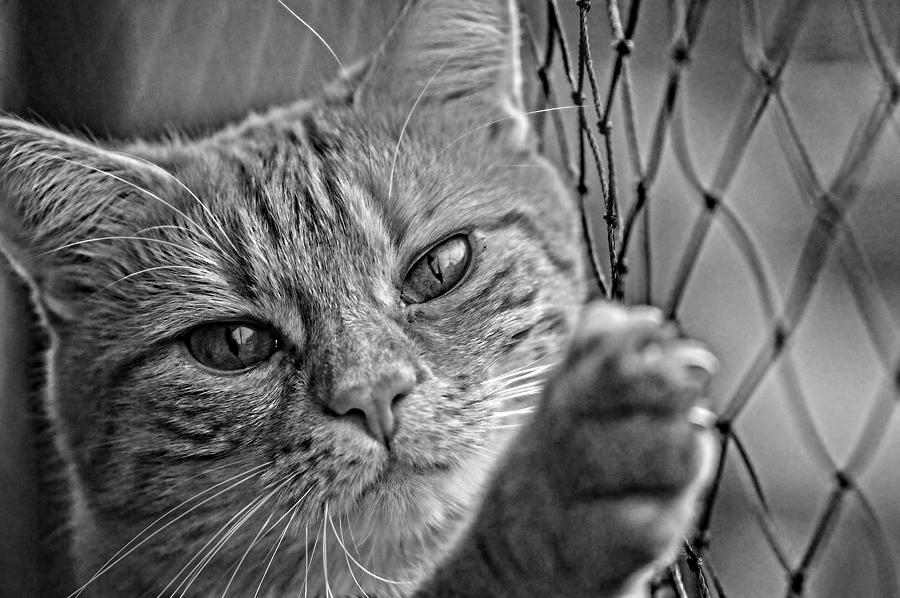}
			\caption{Small Details Enhancement\\($B + 2\times$D1 + $4\times$D2 + $2\times$D3)}
			\label{fig_cat_e1}			
		\end{subfigure}
		\begin{subfigure}[t]{0.325\linewidth}
			\centering
			\includegraphics[width=1.0\linewidth]{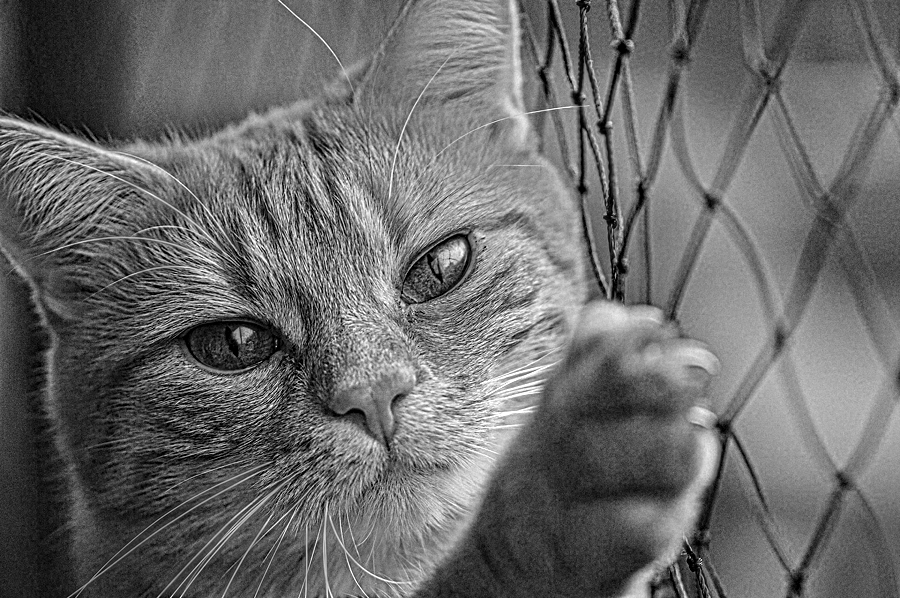}
			\caption{Fine Details Enhancement\\($B + 5\times$D1 + $1.5\times$D2 + $1.25\times$D3)}
			\label{fig_cat_e2}			
		\end{subfigure}
		\begin{subfigure}[t]{0.325\linewidth}
			\centering
			\includegraphics[width=1.0\linewidth]{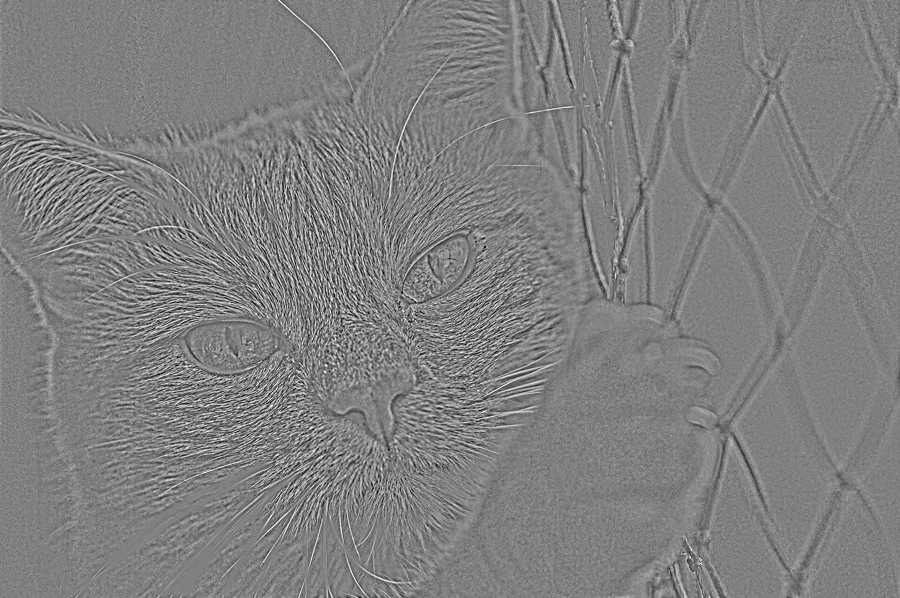}
			\caption{1st detail layer (D1)\\Fine-scale Details}
			\label{fig_cat_d1}			
		\end{subfigure}
		\begin{subfigure}[t]{0.325\linewidth}
			\centering
			\includegraphics[width=1.0\linewidth]{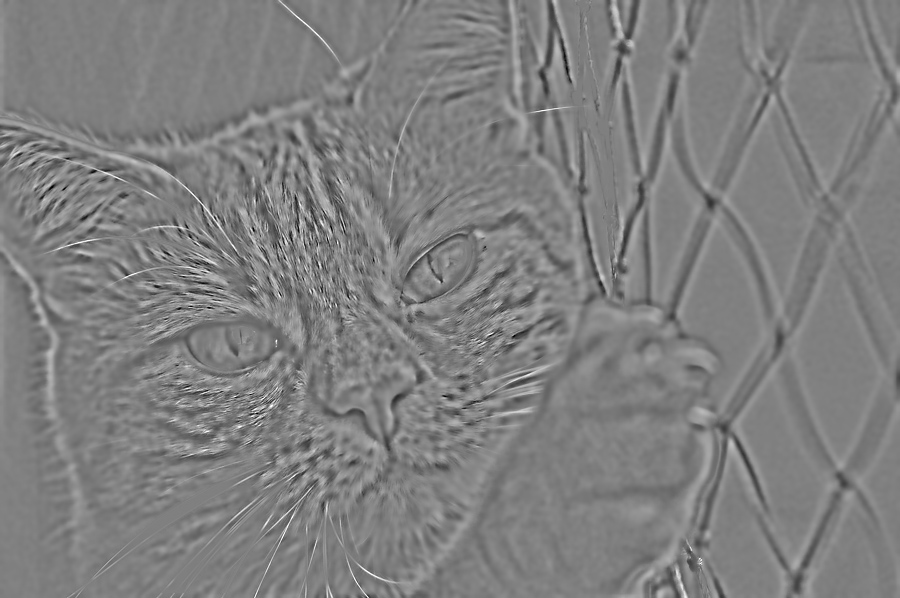}
			\caption{2nd detail layer (D2)\\Small-scale Details}
			\label{fig_cat_d2}
		\end{subfigure}
		\begin{subfigure}[t]{0.325\linewidth}
			\centering
			\includegraphics[width=1.0\linewidth]{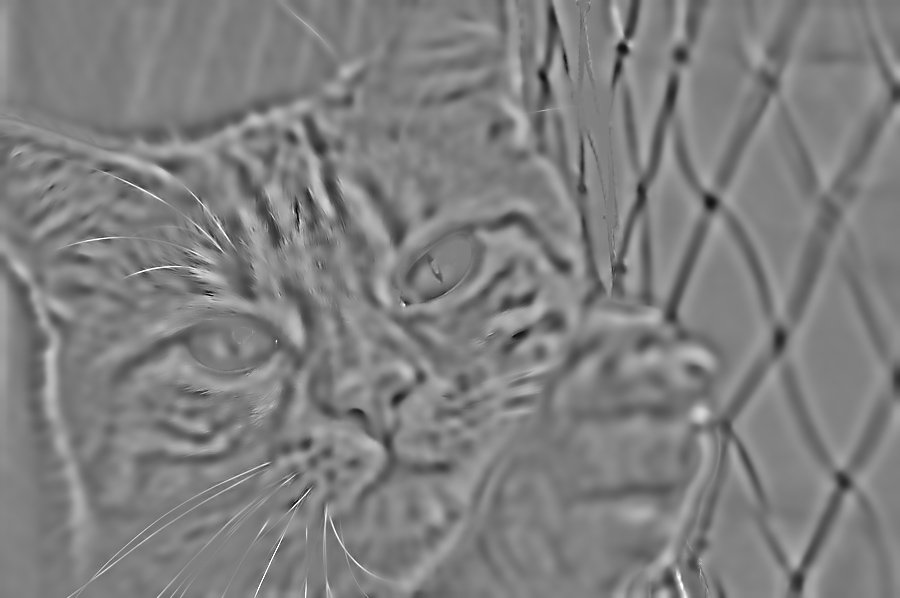}
			\caption{3rd detail layer (D3)\\Medium-scale Details}
			\label{fig_cat_d3}
		\end{subfigure}
		\caption{multi-scale image decomposition using the sub-window variance filter\\3 iterations with $radius = \{2,4,8\}$ and $\epsilon = 0.015$}
		\label{fig_cat}
		
	\end{center}
\end{figure*}

\section{Multi-scale image decomposition}
Applying our edge-preserving filter for multi-scale image decomposition is straight forward. We follow the practice used in the WLS method \cite{farbman2008edge} which is modelled after the Laplacian pyramid \cite{burt1983laplacian}.  The multi-scale decomposition is achieved by decomposing the base image iteratively.  If we write the original image $I$ as $B_0$ with the base and detail layers of the $i^{th}$ level decomposition as $B_i$ and $D_i$, we can show that
\begin{equation}
B_k = B_{k+1} + D_{k+1}
\label{eqn_decomp_04}
\end{equation}
and a multi-scale decomposition with N detail layers can be written as
\begin{equation}
I = B_{N} + \sum_{i=1}^{N}D_i
\label{eqn_decomp_05}
\end{equation}
with
\begin{equation}
B_{k} = E(B_{k-1})
\label{eqn_decomp_06}
\end{equation}
and
\begin{equation}
D_{k} = B_{k-1} - B_{k}
\label{eqn_decomp_07}
\end{equation}
where $E()$ is an edge-preserving filter.

\begin{figure*}
	\begin{center}	
		\begin{subfigure}[t]{0.24\linewidth}
			\centering
			\includegraphics[width=1.0\linewidth]{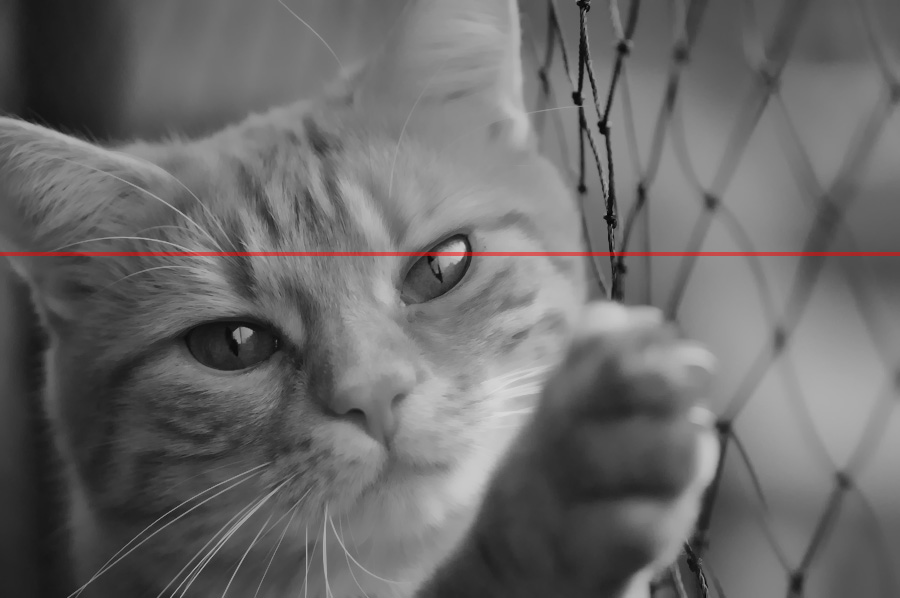}
			\caption{WLS base layer\\$\alpha=1.2, \lambda=0.1$}
			\label{fig_cat1D_WLS_B}
		\end{subfigure}		
		\begin{subfigure}[t]{0.24\linewidth}
			\centering
			\includegraphics[trim={3cm 9.7cm 3cm 10cm},clip,width=1.0\linewidth]{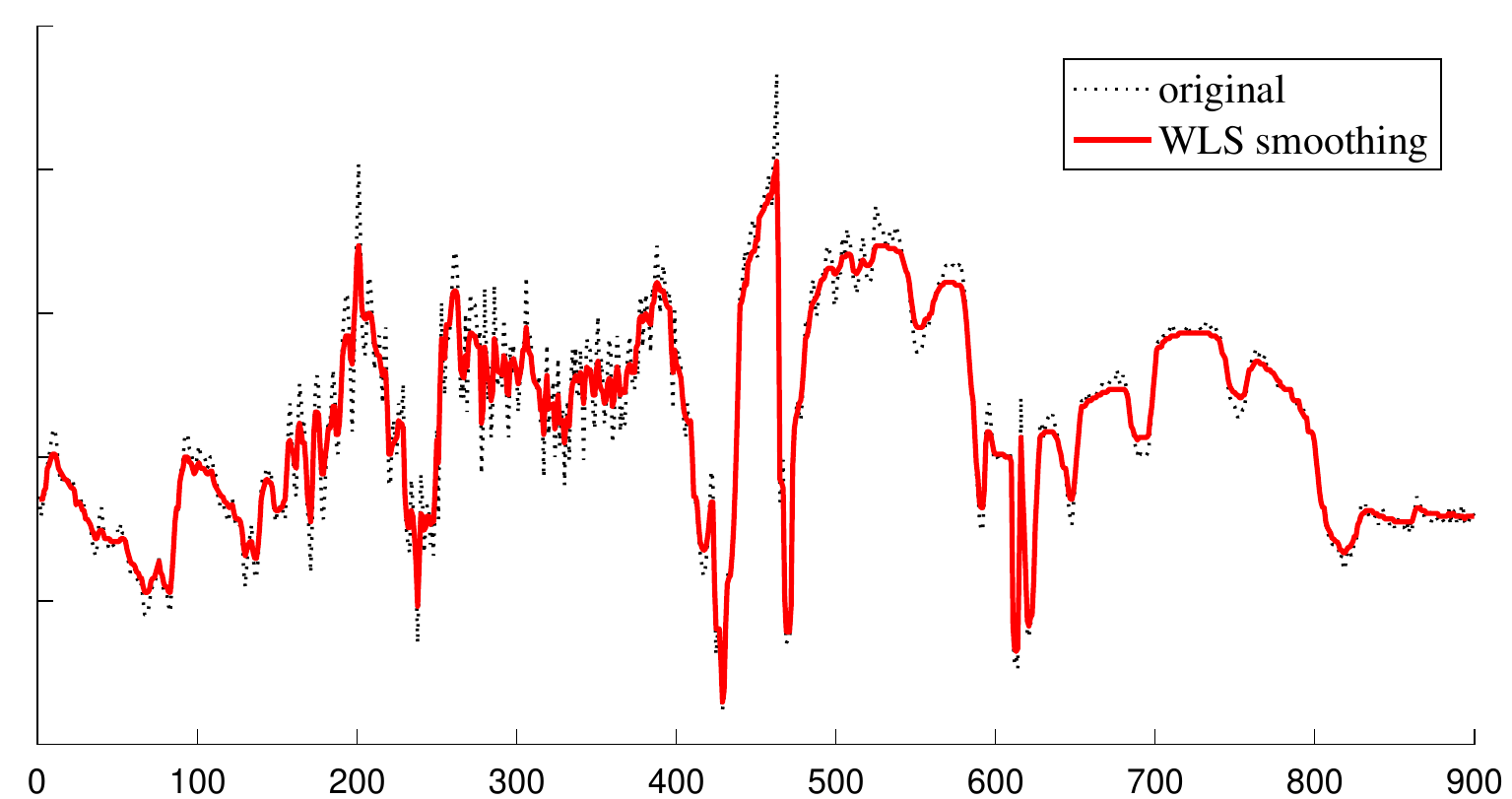}
			\caption{WLS base layer scan-line}
			\label{fig_cat1D_WLS_Bplot}
		\end{subfigure}
		\begin{subfigure}[t]{0.24\linewidth}
			\centering
			\includegraphics[width=1.0\linewidth]{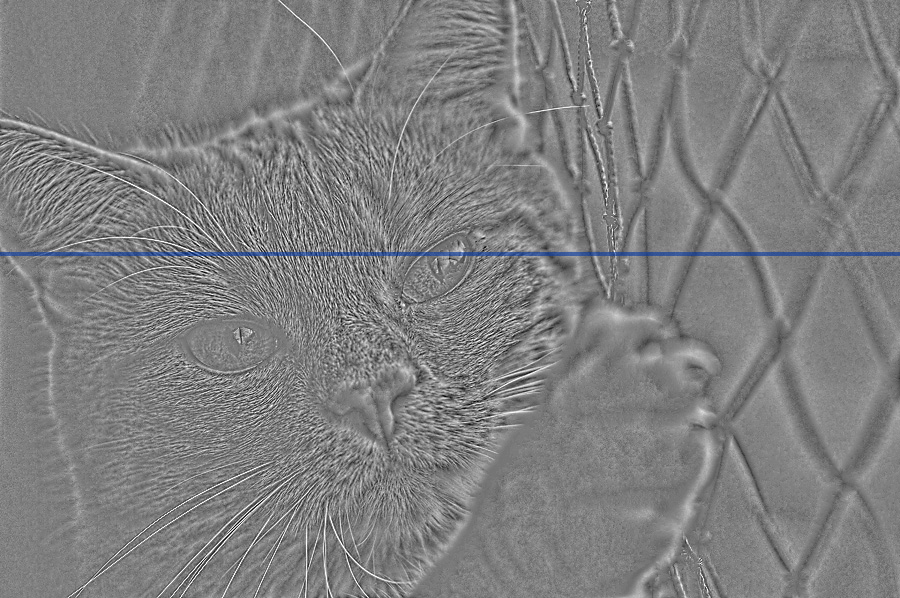}
			\caption{WLS detail layer}
			\label{fig_cat1D_WLS_D}
		\end{subfigure}		
		\begin{subfigure}[t]{0.24\linewidth}
			\centering
			\includegraphics[trim={3cm 9.7cm 3cm 10cm},clip,width=1.0\linewidth]{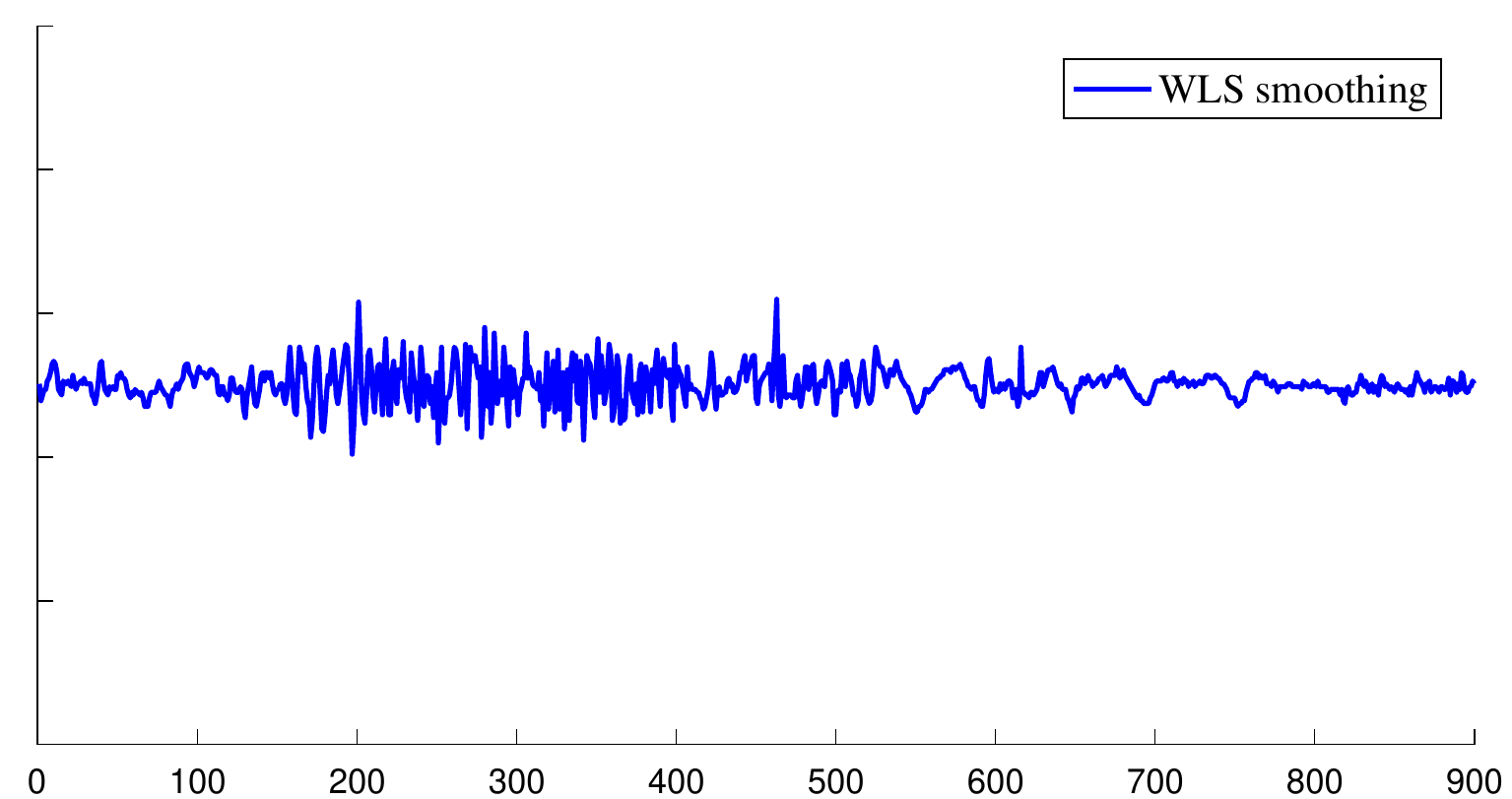}
			\caption{WLS detail layer signal}
			\label{fig_cat1D_WLS_Dplot}			
		\end{subfigure}		
		\vspace{0.2cm}
		
		\begin{subfigure}[t]{0.24\linewidth}
			\centering
			\includegraphics[width=1.0\linewidth]{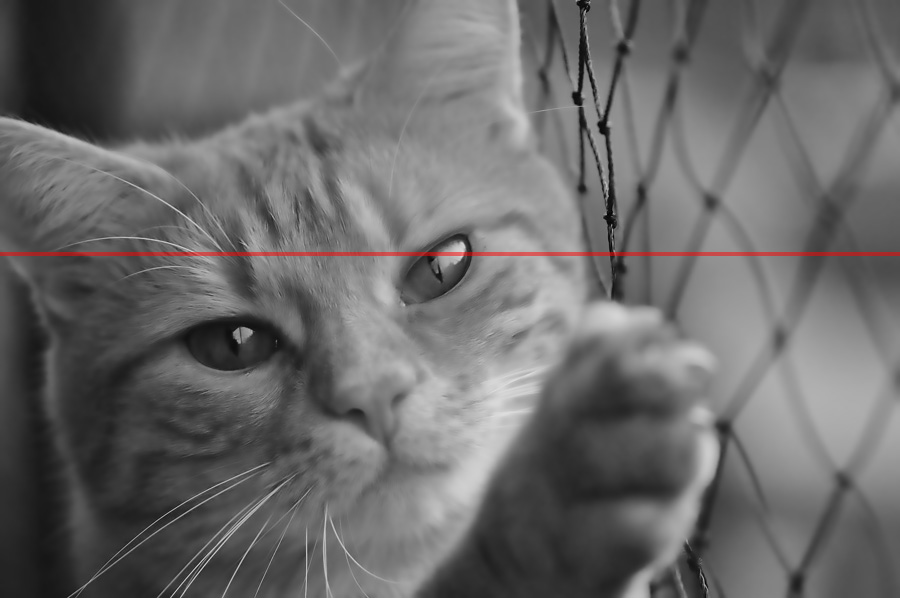}
			\caption{Our base layer\\$radius=2,\epsilon=0.015$}
			\label{fig_cat1D_SVF_B}
		\end{subfigure}
		\begin{subfigure}[t]{0.24\linewidth}
			\centering
			\includegraphics[trim={3cm 9.7cm 3cm 10cm},clip,width=1.0\linewidth]{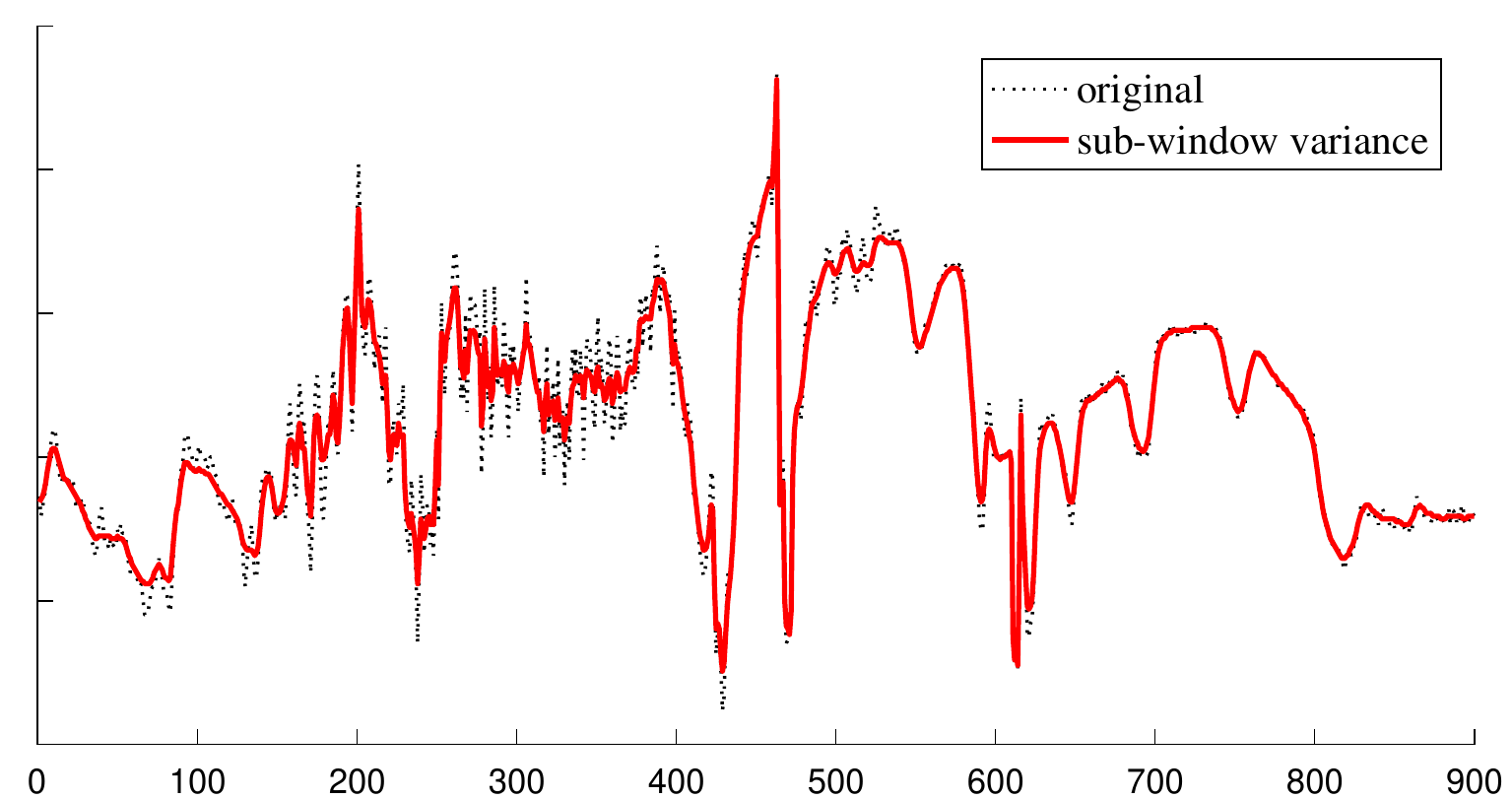}
			\caption{Our base layer scan-line}
			\label{fig_cat1D_SVF_Bplot}			
		\end{subfigure}
		\begin{subfigure}[t]{0.24\linewidth}
			\centering
			\includegraphics[width=1.0\linewidth]{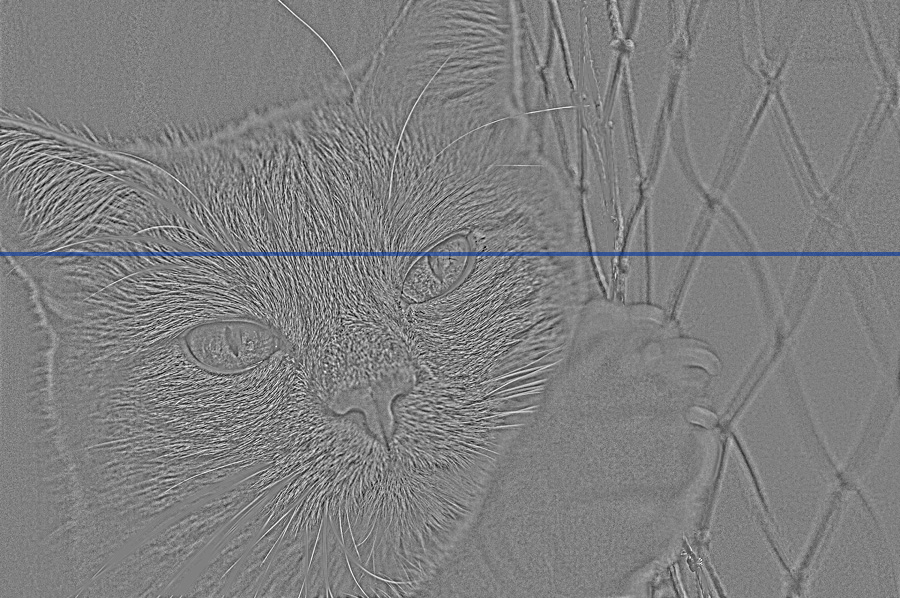}
			\caption{Our detail layer}
			\label{fig_cat1D_SVF_D}
		\end{subfigure}
		\begin{subfigure}[t]{0.24\linewidth}
			\centering
			\includegraphics[trim={3cm 9.7cm 3cm 10cm},clip,width=1.0\linewidth]{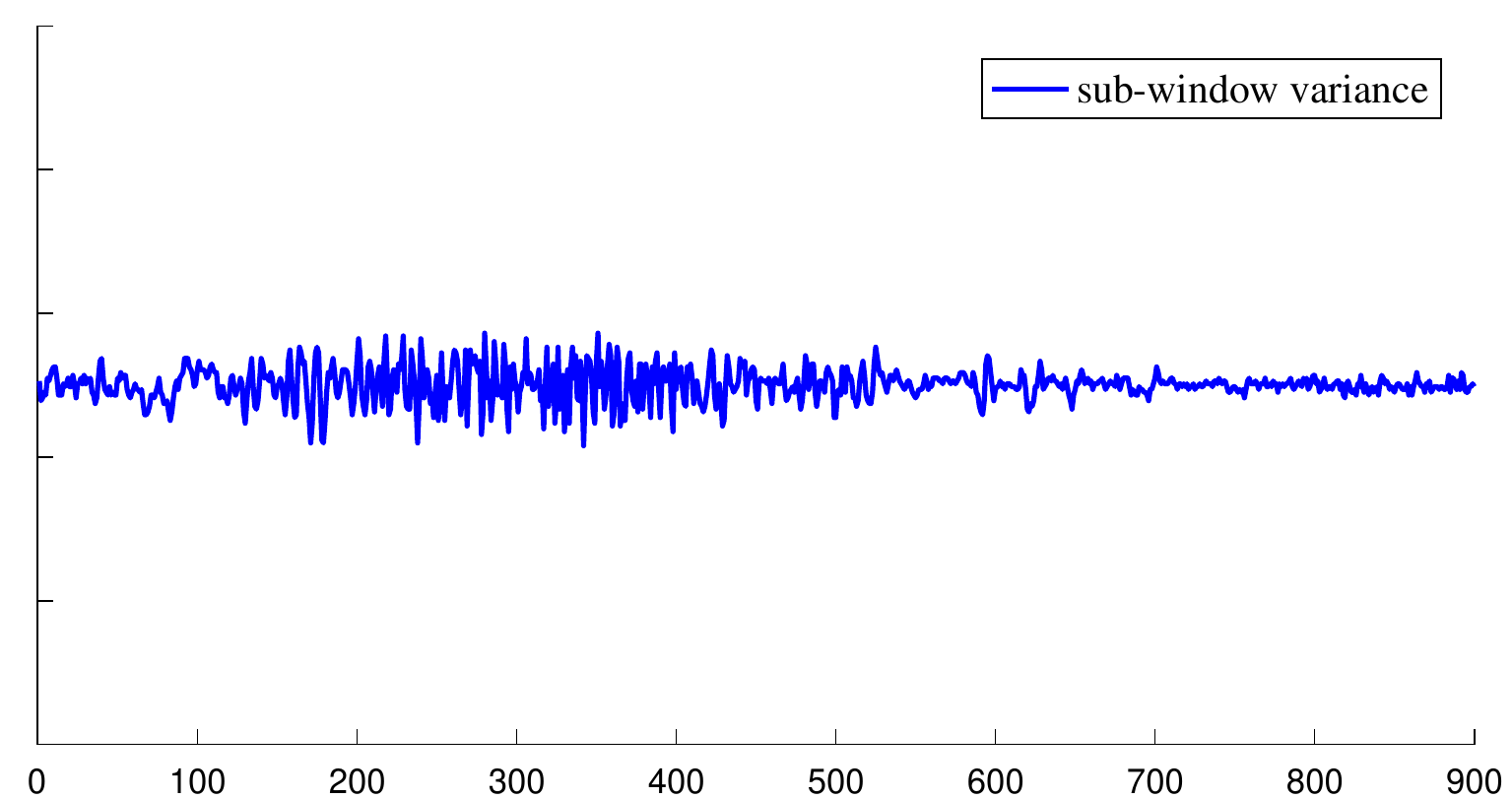}
			\caption{Our Detail layer signal}
			\label{fig_cat1D_SVF_Dplot}
		\end{subfigure}
		\caption{Decomposition difference between the WLS approach and our method}
		\label{fig_cat1D}
	\end{center}
	\vspace{-0.2cm}
\end{figure*}

In each iteration, the filter radius and the edge threshold parameter  $\epsilon$ may be adjusted in a progressive fashion in order to capture the desired features.  We apply our decomposition on an image as shown in Fig.~\ref{fig_cat_org}, it is a four-scale decomposition using a constant $\epsilon=0.015$ and a doubled filter support in each step with an initial filter radius of 2. Strong edges are consistently preserved in each iteration, and our detail layers exhibit uniform similar-scale features and variations.  Fig.~\ref{fig_cat_e1} shows an enhancement where the small-scale details (Fig.~\ref{fig_cat_d2}) is emphasized, and Fig.~\ref{fig_cat_e2} focuses on fine detail (Fig.~\ref{fig_cat_d1}) enhancement. 

\subsection{Importance of Uniformity in the Detail Layers}
The primary goal of multi-scale image decomposition is to enable flexible manipulation of the details of different scales, and it is especially useful to consider \emph{signals of similar spatial-scale and variations} together.  An alternative method \cite{subr2009edge} proposed to group all small spatial-scale signals together regardless of their signal amplitudes.  The disadvantage of such method is that it directly reduces the headroom of possible contrast enhancement for the details which have small amplitude of variations.

Fig.~\ref{fig_cat1D} shows a detailed comparison between our approach and the WLS method.  We tried to extract a fine detail only layer using the WLS method but the best result we obtained still contains details of different scales, and is shown in Fig.~\ref{fig_cat1D_WLS_D}.  This WLS extracted detail layer contains both smooth gradient in the background, and structural details such as the eye's bright highlight.  It is further visualized by the scan-line plot in Fig.~\ref{fig_cat1D_WLS_Dplot}.  

\begin{figure}
	\begin{subfigure}[t]{.49\linewidth}
		\centering
		\includegraphics[width=1.0\linewidth]{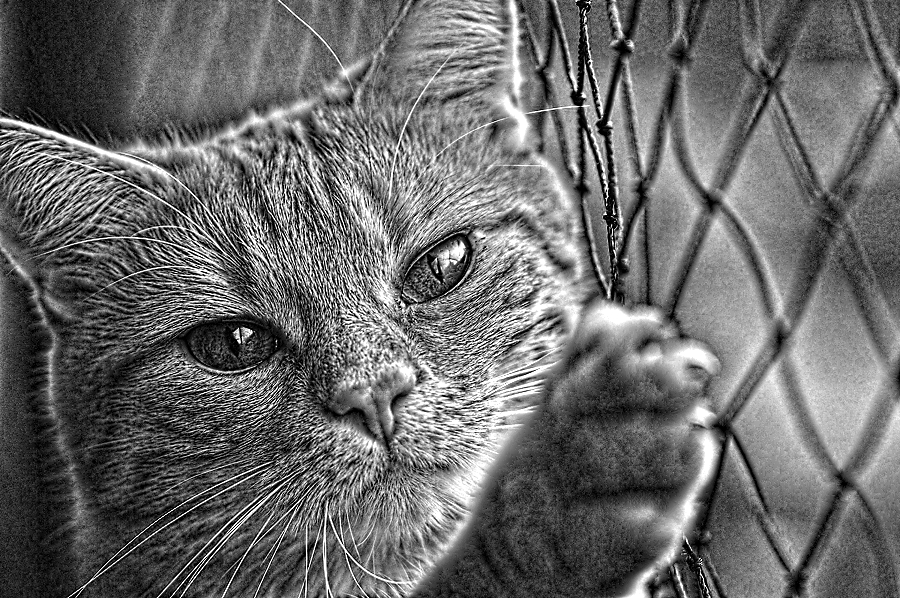}
		\caption{$10\times$ detail enhancement using the WLS approach.}
		\label{fig_cat_WLS_E}
	\end{subfigure}
	\hfill
	\begin{subfigure}[t]{0.49\linewidth}
		\centering
		\includegraphics[width=1.0\linewidth]{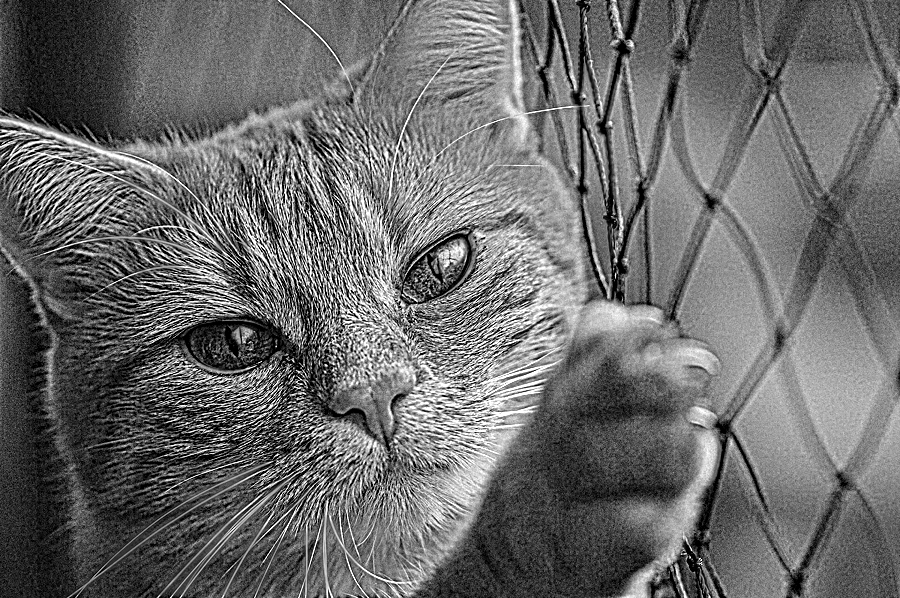}
		\caption{$10\times$ detail enhancement using our approach}
		\label{fig_cat_SVF_E}
	\end{subfigure}		
	\caption{Quality comparison between the WLS approach and our method in an aggressive detail enhancement.}
	\label{fig_cat_QC}
\end{figure}

Figures \ref{fig_cat1D_SVF_B} and \ref{fig_cat1D_SVF_Bplot} show the base layer obtained using our method.  The eye's bright highlight is well preserved, and our detail layer (Fig.~\ref{fig_cat1D_SVF_D}) contains uniformly high frequency details only.  Fig.~\ref{fig_cat_QC} shows a qualitative comparison between the two decomposition methods.  The common goal is to enhance aggressively (a $10\times$ boost of details) the fur details of a cat's picture (original shown in Fig. \ref{fig_cat_org}) using the decomposition results shown in Fig.~\ref{fig_cat1D}.  The WLS enhancement result shown in Fig.~\ref{fig_cat_WLS_E} shows that the global contrast of the whole picture is altered, and many regions have reached peak brightness or blackened, and the whiskers become aliased. 

Our enhancement result as shown in Fig. \ref{fig_cat_SVF_E} exhibits a uniform enhancement of the fur details without affecting the global image contrast.
\\

\section{Applications and Discussions}
\subsection{Interactive Multi-scale Detail Manipulation}
Unlike most previous methods which require pre-computation of the decomposition; our method allows the decomposition stage to run at interactive rate.  We have implemented a simple GPU application using the GLSL shading language, and this application runs on a modest nVIDIA GTX760 GPU.  It is able to decompose one single 1024$\times$1024 RGB floating point image in 20ms using our method, and a three-scale decomposition can run at interactive rate.  We believe that a multi-scale image detail manipulation process is often based on subjective creative decisions, so the possibility to adjust all the decomposition parameters interactively should offer a good manipulation experience.
\\

\begin{figure}
	\begin{center}	
		\begin{subfigure}[t]{1.0\linewidth}
			\centering
			\includegraphics[width=1.0\linewidth]{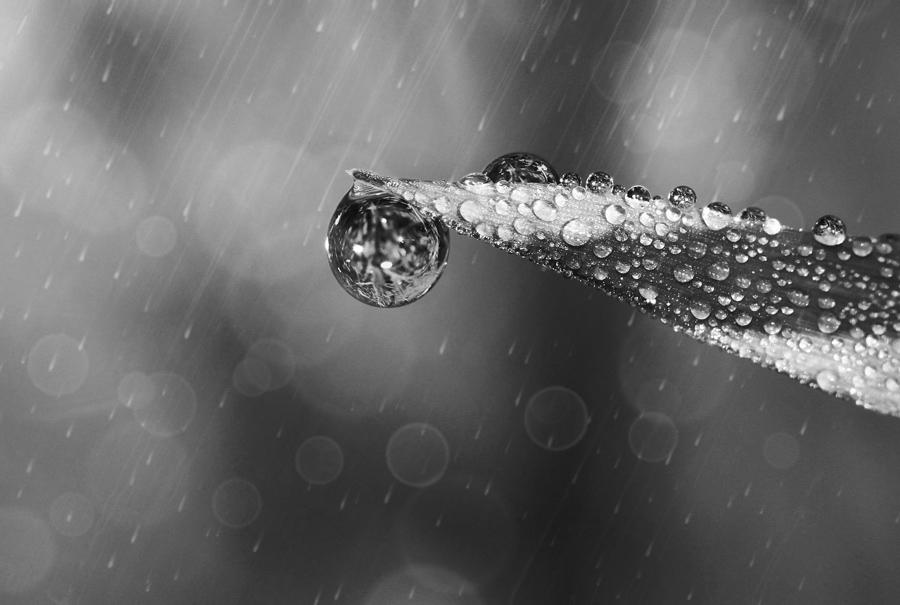}
			\caption{Original image}
			\label{fig_hp_org}			
		\end{subfigure}
		\begin{subfigure}[t]{1.0\linewidth}
			\centering
			\includegraphics[width=1.0\linewidth]{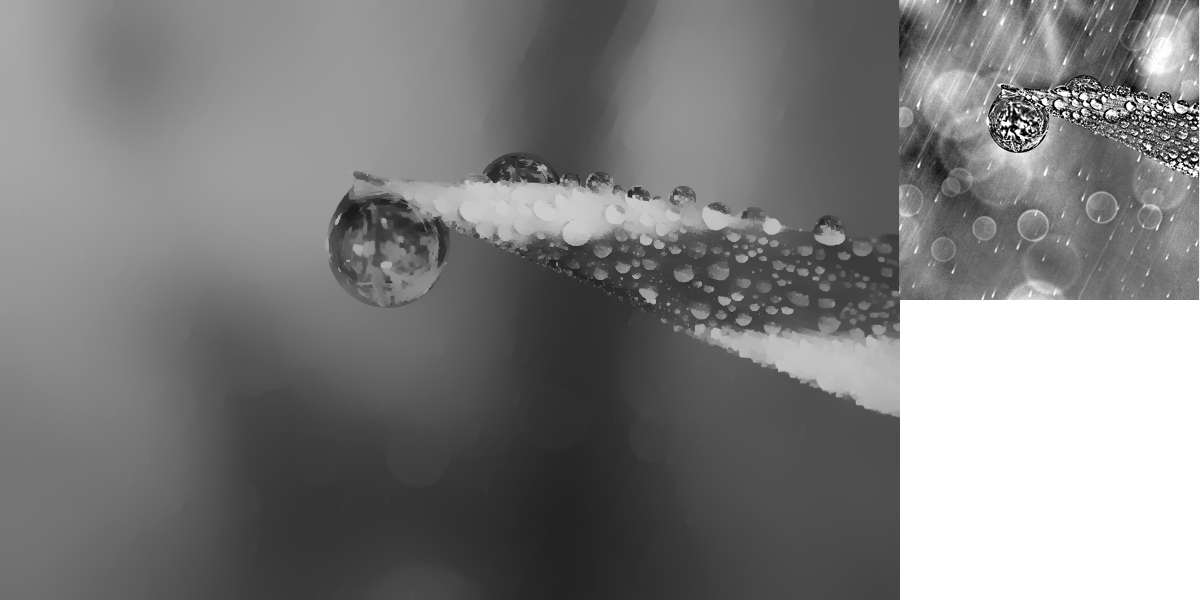}
			\caption{WLS suppression result\\$\alpha=2.0, \lambda=0.4$}
			\label{fig_hp_wls_b}			
		\end{subfigure}		
		\begin{subfigure}[t]{1.0\linewidth}
			\centering
			\includegraphics[width=1.0\linewidth]{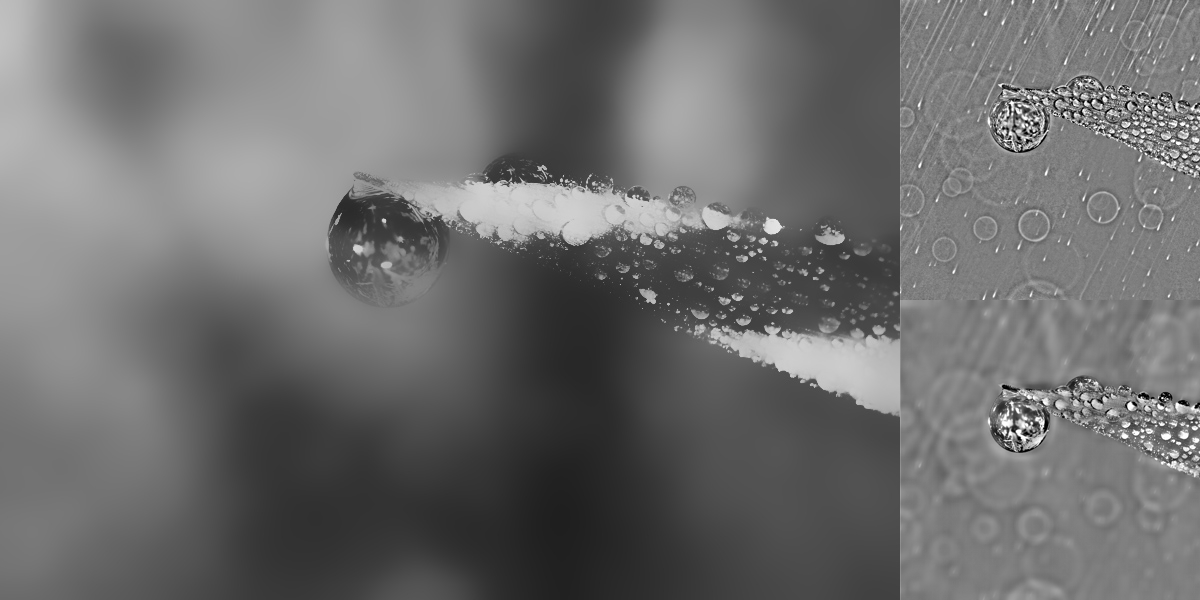}
			\caption{The bilateral filter suppression result\\$\sigma_d=\{10,20\}, \sigma_r = \sqrt{0.03}$}
			\label{fig_hp_bf_base}			
		\end{subfigure}		
		\begin{subfigure}[t]{1.0\linewidth}
			\centering
			\includegraphics[width=1.0\linewidth]{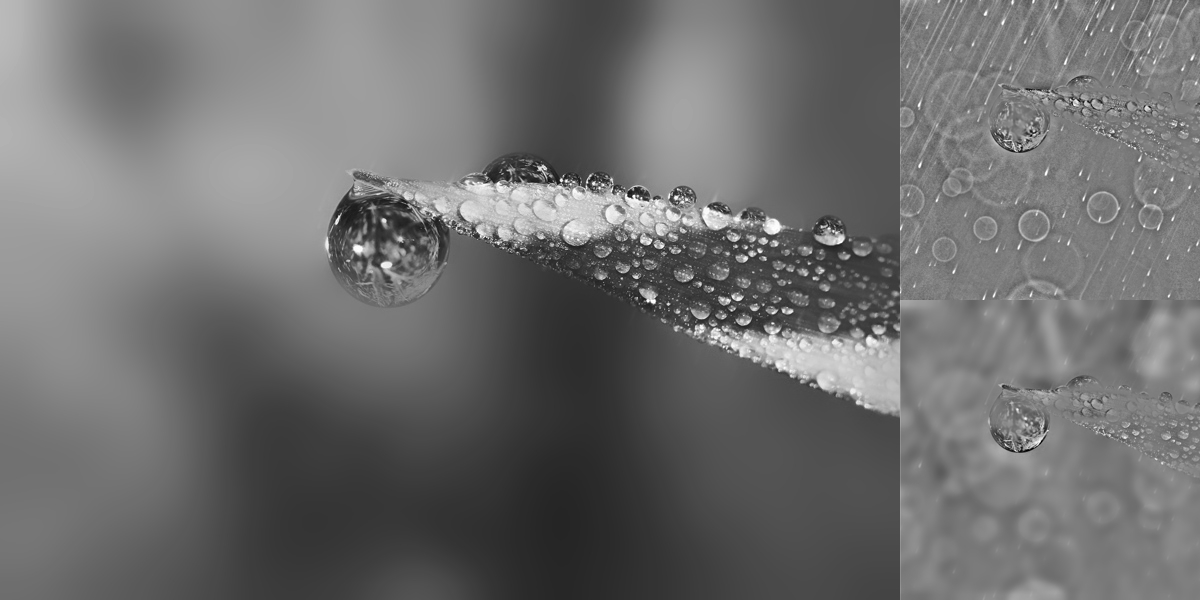}
			\caption{Our suppression result\\$radius=\{10,20\}, \epsilon = 0.03$}
			\label{fig_hp_svf_base}			
		\end{subfigure}
		\caption{Uniform multi-scale details suppression}
		\label{fig_hp}
	\end{center}
\end{figure}

\subsection{Uniform Image Details Suppression}

The ability to extract uniformly same-scale details by our method becomes an important feature which allows us to handle some difficult detail extraction problem as shown in Fig.~\ref{fig_hp}.  The image shown in Fig.~\ref{fig_hp_org} possesses details of different scales, and our goal is to remove the bokehs and the raindrop trails while keeping all other elements intact.

The WLS method's approach makes it difficult to relate its filter's parameters with this task.  It is almost impossible to isolate the target details from the main subject.  We attempted to apply iterative filtering tactics filter but it delivered no improved results.  Fig.~\ref{fig_hp_wls_b} shows its best result with our best effort.  The detail layer (the block on the right hand side of Fig.~\ref{fig_hp_wls_b}) extracted by the WLS approach shows that under the WLS filtering logic, it is fairly difficult to isolate uniformly the details of similar scales.  As a result, its suppression becomes blurry and abstracted.

We apply the bilateral filter to this task, and we use iterated filtering in order to obtain the two desired detail layers.  The filter support size is chosen to remove the thin and small features.  Unfortunately, the detail layers extracted (the blocks on the right side of Fig. \ref{fig_hp_bf_base}) by the bilateral filter contain details of both scales, and we believe it is a consequence of the filter's edge-preservation strategy.  The suppression result (Fig. \ref{fig_hp_bf_base}) by the bilateral filter looks abstracted as expected.

Our filter delivers satisfactory results.  The two detail layers (the blocks on the right side of Fig. \ref{fig_hp_svf_base})) extracted possess uniform variations and share similar spatial scales.  The final base layer (Fig. \ref{fig_hp_svf_base}) also fulfils the goal, and the result is significantly better than the other two leading methods.  The high sensitivity of structure by the sub-window variance filter enables a lot of fine edge-like features to be well preserved on the base layer.  This uniform extraction capability makes our proposed method particularly suitable for refined detail manipulation tasks which demand separation of details.

\subsection{Limitations}
There exist certain situations where the users may want to extract or suppress image details of a limited range of tone only.  Our current formulation preserves edge-alike features based on a constant threshold variance value, i.e. the parameter $\epsilon$.  This tone selective suppression may become challenging to our method, one possibility is to make the parameter $\epsilon$ tone adaptive, such as to extend this parameter into a function of mean pixel intensity of the filter window.
\\

\section{Conclusion}
We have presented a multi-scale image decomposition approach based on a novel local statistical edge model which delivers interactive performance.  Our method emphasizes the quality of edge preservation, and the importance of uniformity of details in terms of both spatial scale and variations during decomposition. The simplicity of our intuitive method is based on an important observation of edge properties from the perspective of perception and simple image statistics.
\\


\bibliographystyle{IEEEtran}
\bibliography{icvr_2021_v1} 

\end{document}